\documentclass[10pt,twocolumn,letterpaper]{article}
\usepackage[pagenumbers]{iccv}
\usepackage{booktabs}
\usepackage{arydshln}
\usepackage{makecell}
\usepackage{enumitem}
\usepackage{amssymb}
\usepackage{graphicx}
\usepackage{soul}
\usepackage{url}
\usepackage{amsmath}
\usepackage{amsthm}
\usepackage{algorithm}
\usepackage{algorithmic}
\usepackage{multirow}
\usepackage{marvosym}
\usepackage{threeparttable}
\usepackage{mathrsfs}
\usepackage{color,xcolor}
\usepackage{colortbl}
\definecolor{iccvblue}{rgb}{0.21,0.49,0.74}
\usepackage[pagebackref,breaklinks,colorlinks,allcolors=iccvblue]{hyperref}
\def\paperID{672}
\def\confName{ICCV}
\def\confYear{2025}

\usepackage{tocbibind}
\usepackage{paralist}

\begin{document}
\title{Vector Contrastive Learning For \emph{Pixel-Wise} Pretraining In Medical Vision}

\author{Yuting He$^{1}$ \quad Shuo Li$^{1,2,}$\thanks{Corresponding author: shuo.li11@case.edu}\\
$^{1}$ Dept. of BME, $^{2}$ Dept. of CDS, Case Western Reserve University, US, Cleveland\\
{\tt\small \{yuting.he4, shuo.li11\}@case.edu}\\
{\small code: \url{https://github.com/YutingHe-list/COVER}}
}

\maketitle

\begin{abstract}
Contrastive learning (CL) has become a cornerstone of self-supervised pretraining (SSP) in foundation models; however, extending CL to pixel-wise representation—crucial for medical vision—remains an open problem. Standard CL formulates SSP as a binary optimization problem (binary CL) where the excessive pursuit of feature dispersion leads to an ``over-dispersion" problem, breaking pixel-wise feature correlation thus disrupting the intra-class distribution. Our vector CL reformulates CL as a vector regression problem, enabling dispersion quantification in pixel-wise pretraining via modeling feature distances in regressing displacement vectors. To implement this novel paradigm, we propose the \textbf{CO}ntrast in \textbf{VE}ctor \textbf{R}egression (\textbf{COVER}) framework. COVER establishes an extendable vector-based self-learning, enforces a consistent optimization flow from vector regression to distance modeling, and leverages a vector pyramid architecture for granularity adaptation, thus preserving pixel-wise feature correlations in SSP. Extensive experiments across 8 tasks, spanning 2 dimensions and 4 modalities, show that COVER significantly improves pixel-wise SSP, advancing generalizable medical visual foundation models.

\end{abstract}
\section{Introduction}\label{sec:intro}
Contrastive learning (CL) \cite{tian2020makes,Chen2021CVPR,chen2020simple,oquab2024dinov2,grill2020bootstrap,caron2018deep,chen2020improved} has become a cornerstone of self-supervised pretraining (SSP) \cite{mannaself} in medical visual foundation models (MVFMs) \cite{wu2024large,chen2024towards,vorontsov2024foundation,zhou2023foundation,moor2023foundation,he2024foundation}. However, applying CL in a \textit{pixel-wise} manner, an approach pivotal for disentangling the clinically critical semantics within medical images \cite{zhang2024single}, remains an open problem. Current preliminary works \cite{wu2024large,chen2024towards,vorontsov2024foundation} adopt image-wise contrast strategies derived from natural image studies \cite{chen2020simple,chen2020improved,oquab2024dinov2,Chen2021CVPR,grill2020bootstrap}, which emphasize global representations \cite{wang2022densecl}. In medical images, the global context is often similar due to the consistent anatomies in human body \cite{netter2014atlas}, leading to a lack of global diversity that will cause CL models to converge to trivial solutions \cite{saunshi2019theoretical,wang2020understanding}. Recent evidence highlights the critical importance of local features \cite{xie2021propagate,li2021dense,wang2022densecl,zhang2024single} and leading results \cite{haghighi2024self,yan2022sam,quan2024igu} in medical images increasingly incorporate pixel-wise methods. This observation raises a fundamental question: \textit{Can CL be reformulated as a truly pixel-wise SSP for MVFMs?}

\begin{figure}
    \centering
    \includegraphics[width=1\linewidth]{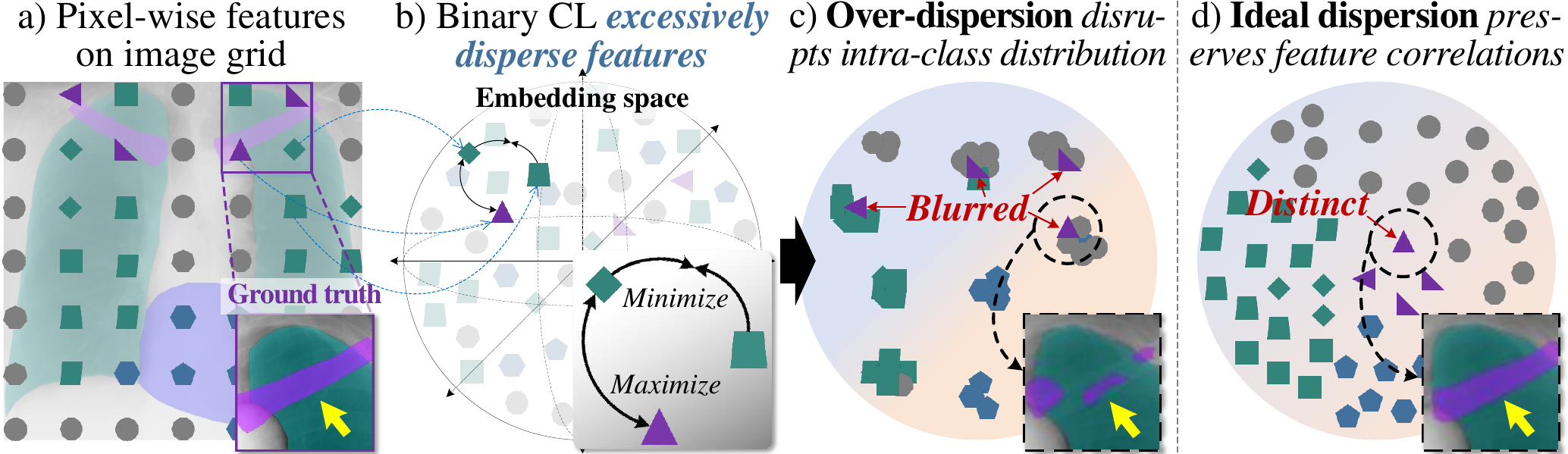}
    \caption{\textbf{Over-dispersion problem}: Binary CL excessively pursues to disperse features, disrupting intra-class distribution and struggling models to disentangle the underlying semantics.}
    \label{fig:1}
\end{figure}
An obstacle to answering this question is the \emph{over-dispersion problem} \cite{hinde1998overdispersion,wang2020understanding,saunshi2019theoretical}. The standard CL formulates the SSP as a binary optimization problem (termed \emph{binary CL}), where feature distances in positive or negative pairs are directly minimized or maximized \cite{oord2018representation}, leading to excessive dispersion of features without degree modeling (Fig.\ref{fig:1}-b). However, recent findings indicate that excessive pursuit of dispersion will disrupt intra-class distribution \cite{wang2020understanding,wang2021understanding}. This rigid contrast imposes significant limitations on medical vision tasks that require high sensitivity to subtle variations and structures, like thin vessels and subtle lesions, which risk being blurred or indistinguishable (Fig.\ref{fig:1}-c). While some attempts \cite{wu2024voco,chaitanya2020contrastive} softened the contrast or introduced regularization \cite{wang2020understanding} to mitigate this problem, they remain fundamentally reliant on the binary CL.

In pixel-wise pretraining, the over-dispersion problem renders the binary CL particularly ineffective (Fig.\ref{fig:1} a-c). Pixel-wise features are distributed across the image grid, where semantics vary continuously \cite{gonzales1987digital}, making them inherently correlated. Unlike discrete classification tasks, continuous signals lack a natural definition of distinct categories \cite{he2020momentum,thoma2020soft}, leading to ambiguity near proximity divisions. The excessive pursuit of dispersion breaks these correlations so that it aggravates the disruption of the inherent intra-class distribution, struggling pretrained models to disentangle the underlying semantics hidden in pixels \cite{bengio2013representation}.

\textit{Motivation:} Modeling distances for pixel-wise features is essential to quantify their dispersion degrees. The binary CL follows a max-min contrast for feature distances, neglecting explicit quantification of dispersion (Fig.\ref{fig:1}-b). Ideally, suppose we can predefine a ground truth distance $\alpha$ to indicate the desired feature dispersion, the model will be trained to align feature distances accordingly \cite{yang2006distance,suarez2021tutorial}. Such a CL with explicit distance modeling would enable precise quantification of dispersion while preserving feature correlations (Fig.\ref{fig:1}-d). Unfortunately, in practice, these ground truth distances are task-dependent \cite{qu2005new}, making them unmeasurable in the SSP and mismatched with the general objectives of foundation models \cite{bommasani2021opportunities}.

\begin{figure}
    \centering
    \includegraphics[width=\linewidth]{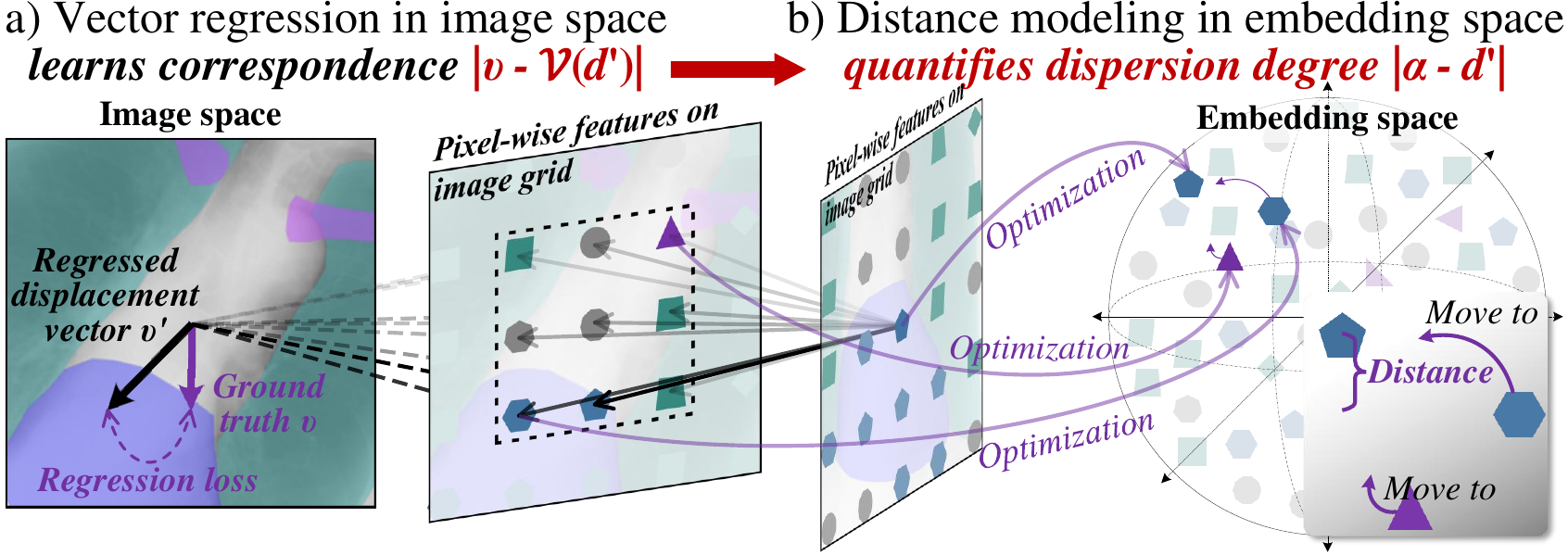}
    \caption{\textbf{Motivation}: Our vector CL unifies the distance modeling $|d-d'|$ and displacement vector regression $|v-\mathcal{V}(d')|$, enabling CL with quantitative dispersion via the optimization flow.}
    \label{fig:2}
\end{figure}

This paper proposes a solution for the over-dispersion problem in pixel-wise CL by formulating a \emph{vector contrastive learning}. The key insight is that feature distances inherently encode semantic correspondences and can be represented as displacement vectors in image space (Fig.\ref{fig:2}). When pixel-wise features align on an image grid, such vectors (pointing from source to target coordinates) provide a structured way to model correspondences. Therefore, instead of directly minimizing $|\alpha-d'|$ to model distances $d'$ in the embedding space, we formulate a function $\mathcal{V}$ that relates distances to vectors $v'$ in the image space. This reformulates the CL as a vector regression problem: $|v-\mathcal{V}(d')|$ (formulated in Sec.\ref{sec:form}), where $v$ is the ground truth vector. By leveraging the properties of vectors, this approach structurally regulates feature dispersion, preserving semantic continuity. Incorporating measurable magnitudes and directions from the image space, vector CL provides a principled way to guide CL with distance modeling.

\begin{figure*}
    \centering
    \includegraphics[width=0.91\linewidth]{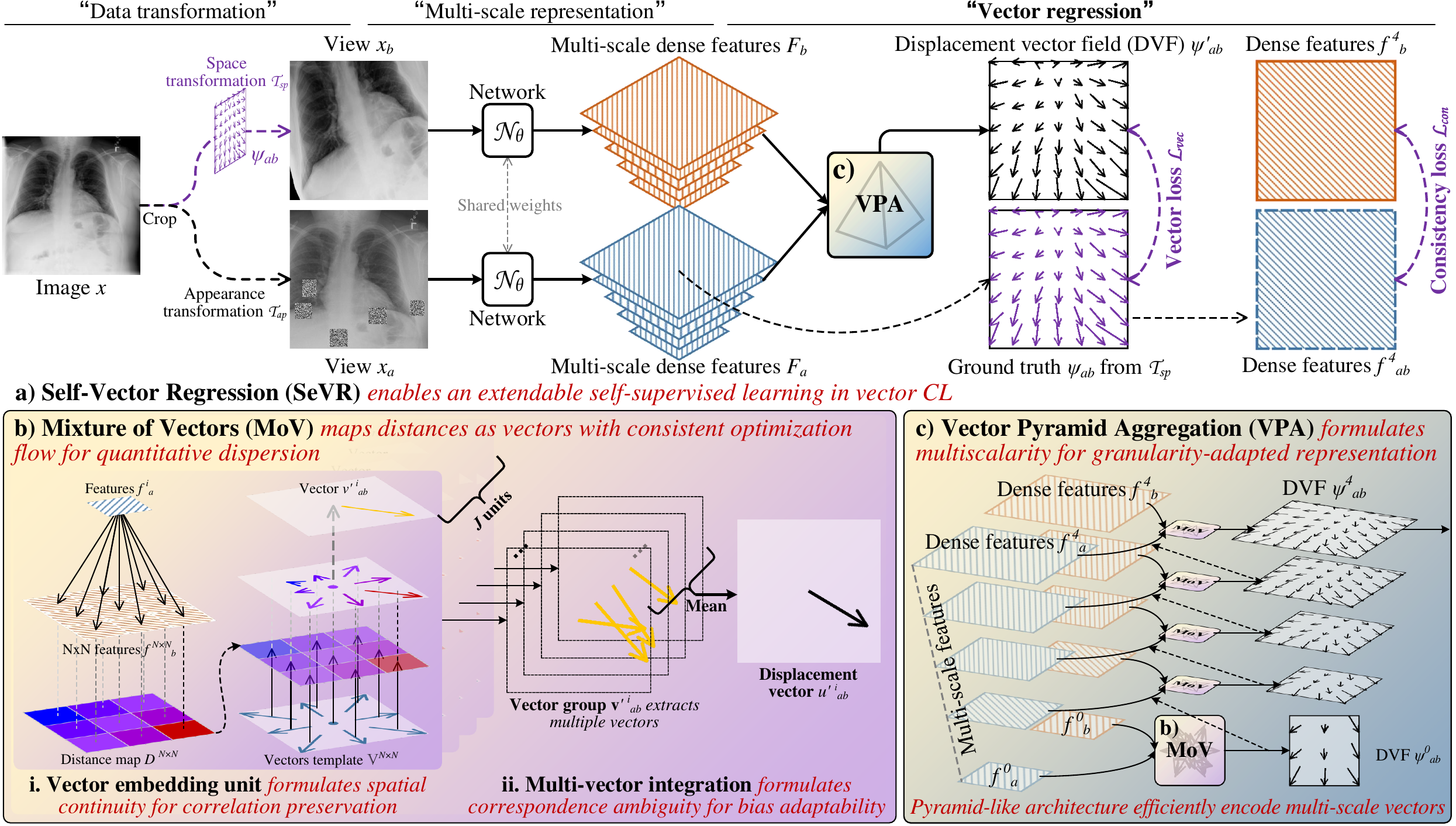}
    \caption{\textbf{Framework:} Our COVER formulates the distance modeling as the regression of displacement vectors (coordinate correspondences), quantifying dispersion degree in pixel-wise CL. a) SeVR (Sec.\ref{subsec:VL}) implements vector CL in a self-space transformation process. b) MoV (Sec.\ref{subsec:mov}) formulates feature distances as vectors with the properties of spatial continuity (VEU (i)) and correspondence ambiguity (MVI (ii)). c) VPA (Sec.\ref{subsec:vpa}) formulates multiscalarity of correspondence via integrating the multi-scale vectors for granularity adaptability.}
    \label{fig:3}
\end{figure*}

The implementation of vector CL remains unexplored, hindered by two questions: Q1. How to construct a self-learning paradigm with free ground truth vector $v$ that can be extended across diverse medical images? Q2. How to formulate the function $\mathcal{V}$ to ensure a consistent optimization flow from vector regression to distance modeling while aligning with the properties of correspondence? Although preliminary studies \cite{He_2023_CVPR,he2025homeomorphismpriorfalsepositive} explored vectors in medical SSP, they lacked a structured formulation to link distances and vectors, causing inconsistent optimization flows. \emph{For the first time}, we formulate the vector CL and implement it as a \textbf{CO}ntrast in \textbf{VE}ctor \textbf{R}egression (\textbf{COVER}) framework, which models distance in pixel-wise CL, enabling quantitative dispersion in pixel-wise SSP via three key innovations: 

\emph{1) Self-Vector Regression (SeVR)} (\emph{for Q1}) constructs an \textit{extendable} self-space transformation mechanism for vector CL with free ground truth vectors $v$ (Fig.\ref{fig:3}-a, Sec.\ref{subsec:VL}). Specifically, it generates a displacement vector field (DVF), a dense field of vectors on the image grid, capturing pixel-wise correspondences between two transformed views of a medical image. The pixel-wise distances are modeled through vector regression supervised by the DVF. This synthesis-based process enables a scalable and annotation-free SSP algorithm, facilitating vector CL across a wide variety of medical images (free from paired images \cite{He_2023_CVPR,he2025homeomorphismpriorfalsepositive}).

\emph{2) Mixture of Vectors (MoV)} (\emph{for Q2}) formulates a mapping function with consistent optimization flow from vector regression to distance modeling (Fig.\ref{fig:3}-b, Sec.\ref{subsec:mov}) with two properties: \emph{i. Spatial continuity.} It designs a vector embedding unit (VEU) that encodes continuous spatial relationships in a vector template to map pixel-wise feature distances to vectors, thus avoiding the artificial division \cite{li2021dense,wang2022densecl,xie2021propagate} and preserving the feature correlation. \emph{ii. Correspondence ambiguity.} It constructs a multi-vector integration (MVI) strategy that extracts multiple vectors indicating potential correspondences with diverse feature concerns, enhancing bias adaptability for pretrained models.

\emph{3) Vector Pyramid Aggregation (VPA)} (\emph{for Q2}) formulates \emph{multiscalarity} of correspondence via stacking the MoV in a pyramid-like architecture (Fig.\ref{fig:3}-c, Sec.\ref{subsec:vpa}). It predicts vectors across multiple feature scales and fuses them to capture correspondences for semantic objects at varying scale levels. Therefore, it enables the vector CL on different scales with low computation cost, training the model to represent multiscale features and improving granularity adaptability for pretrained models.

Our contributions are summarized as follows: 1) For the first time, we pioneer a novel CL paradigm, vector CL, overcoming the over-dispersion problem in standard binary CL while preserving intra-class distribution. 2) We propose the COVER framework, which implements the vector CL for an extendable pixel-wise SSP in MVFMs. 3) Our SeVR establishes a versatile self-learning paradigm extendable across diverse medical images. 4) Our MoV maps feature distances as vectors, enabling a consistent optimization flow that quantifies dispersion while maintaining feature correlations. 5) Our VPA captures multiscale correspondences, enhancing representations across varying granularity scales.

\section{Related works}
\textbf{1) Medical image self-supervised pretraining} \cite{mannaself} (SSP), as a training algorithm without annotation, is effectively promoting the programming of medical visual foundation models \cite{he2024foundation,chen2024towards,zhou2023foundation,moor2023foundation}. It will capture the posterior distribution of the underlying explanatory factors and make models easier to extract useful information \cite{bengio2013representation}, thus achieving the general application goal. It always utilizes the generative learning (GL) \cite{vincent2010stacked,pathak2016context,zhou2019models,komodakis2018unsupervised} or contrastive learning (CL) \cite{tian2020makes,Chen2021CVPR,chen2020simple,oquab2024dinov2,grill2020bootstrap,caron2018deep,li2021dense,wang2022densecl,xie2021propagate} paradigms that train model by generating generate meaningful information or contrasting feature pairs. In this paper, we limit our scope to medical visual CL and review the below topics that are relevant to the applications considered in the sequel.

\textbf{2) Binary contrastive learning} has become a dominant SSP paradigm that learns general representations without annotations. Early works \cite{chen2020simple,he2020momentum,grill2020bootstrap} maximize the distance between positive pairs while minimizing distance with negatives, leveraging global representations in natural images and being implemented to medical images \cite{wu2024large,chen2024towards,vorontsov2024foundation}. However, the lack of global diversity in medical images makes the global representations insufficient, causing model collapse \cite{saunshi2019theoretical,wang2020understanding}. While dense binary CL \cite{xie2021propagate,wang2022densecl} learns dense representations for local features, it suffers from \emph{over-dispersion problem} due to mini-max dispersion, distorting their correlations and intra-class distributions \cite{wang2020understanding,wang2021understanding}.

\textbf{3) Distance modeling} \cite{yang2006distance,suarez2021tutorial} quantifies features' distances in embedding spaces, preserving their correlations. It focuses on learning similarity metrics and guiding feature alignment with specific distance \cite{hoffer2015deep,sohn2016improved}. This approach aligns closely with human cognition in perceiving continuous objects \cite{contier2024distributed}. This has proven valuable in capturing feature correlations for effective pixel-wise feature representation. However, when aiming for SSP with general representations, predefined distances become impractical \cite{qu2005new}, posing challenges to direct distance modeling.

\textbf{4) Vector regression} was originally developed to learn correspondence vectors of visual objects, and has been widely applied in image registration \cite{9842340,he2021few,balakrishnan2019voxelmorph,wang2023modet}, geometric matching \cite{rocco2017convolutional,dong2023rethinking,ehm2024partial}, etc. Some preliminary SSP works \cite{zhu2020rubik,komodakis2018unsupervised} attempt to establish absolute correspondence between semantics and positions but overlook their relative variations. GVSL \cite{He_2023_CVPR} and GEMINI \cite{he2025homeomorphismpriorfalsepositive} introduce correspondence learning into medical SSP by predicting vector fields that represent the dense correspondence between objects across images. However, the lack of a formulation for the mapping function between feature distances and vectors has limited its ability to exert control over pixel-wise representations. This paper addresses this gap by formulating the relationship between feature distances and their vectors.
\section{COVER: COntrast in VEctor Regression}\label{sec:method}
Our COVER (Fig.\ref{fig:3}) implement the formulated vector CL (Sec.\ref{sec:form}) via establishing the consistent optimization flow between distance modeling and vector regression (MoV, Sec.\ref{subsec:mov}), modeling multi-scale representation (VPA, Sec\ref{subsec:vpa}), and learning the vector regression in a self-learning pipeline (SeVR, Sec.\ref{subsec:VL}), thus yielding pixel-wise pretraining with quantitative dispersion in medical vision.

\subsection{Formulation}\label{sec:form}
\textbf{3.1.1 Problem formulation} The proposed vector CL reformulated the CL as a vector regression problem that models pixel-wise feature distances to quantify their dispersion degrees. It defines a mapping function $\mathcal{V}$ that links pixel-wise features to displacement vectors with a consistent optimization flow. Given an input medical image $x$ from dataset $\mathcal{D}$, we apply a random spatial transformation function $\mathcal{T}_{sp}$ to generate two augmented views $x_a,x_b$ with their spatial correspondence encoded in a displacement vector field (DVF) $\psi_{ab}=\{v^{i}\}_{i\in\Omega}$ ($\Omega$ is the image grid), where $v^{i}$ represents the coordinate shift from $x_a$ to $x_b$ at pixel $i$. The network $\mathcal{N}_{\theta}$ extracts dense feature maps $F_a=\mathcal{N}_{\theta}(x_a)$ and $F_b=\mathcal{N}_{\theta}(x_b)$. The function $\mathcal{V}$ then predicts the DVF $\psi'_{ab}=\{v'^{i}\}_{i\in\Omega}$ from these features, aiming to minimize the disparity between $\psi'_{ab}$ and $\psi_{ab}$. This is formulated as:
\begin{align}\label{equ:form}
  \min_{\theta}\ & \mathbb{E}_{x\sim\mathcal{D}}[|\psi'_{ab}-\psi_{ab}|],\\
  \text{where}\ & \psi'_{ab}=\mathcal{V}(F_a,F_b),\ \psi_{ab}\sim\mathcal{T}_{sp},\ \psi_{ab}(x_a)\equiv x_b. \notag
\end{align}
The $\equiv$ means both sides have consistent space.

\textbf{3.1.2 Why does our vector CL drive distance modeling?} Because the vector CL unifies feature distance modeling and displacement vector regression mathematically. It constructed an equivalency of vector regression and distance modeling, obtaining a consistent optimization flow, and thus enabling the ability to learn feature distances in the embedding space without ground truth distances. This process is formalized as follows: 
\begin{align}
\emph{\text{Dist. Mod.:}}\ &\begin{matrix}\sum_{i=0}^{I}|\alpha^{i}-d'^{i}|\rightarrow0\end{matrix}\label{equ:dis1}\\
  &\Leftrightarrow \begin{matrix}\sum_{i=0}^{I}\mathbb{V}^{i}|\alpha^{i}-d'^{i}|\rightarrow0\end{matrix}\label{equ:dis2}\\
  &\Rightarrow \begin{matrix}|\sum_{i=0}^{I}\mathbb{V}^{i}\alpha^{i}-\sum_{i=0}^{I}\mathbb{V}^{i}d'^{i}|\rightarrow0\end{matrix}\label{equ:dis3}\\
  \emph{\text{Vec. Reg.:}}\ &\Leftrightarrow\begin{matrix}|v-\sum_{i=0}^{I}\mathbb{V}^{i}d'^{i}|\rightarrow0.\end{matrix}\label{equ:dis4}
\end{align}
The ideal distance modeling in a set can be formulated as Equ.\ref{equ:dis1}, where the $\alpha^{i}$ is the ground truth distance for feature pair $i$ in the set with $I$ distances. Our COVER defines this distance set as the distances on a field of the image grid (Sec.\ref{subsec:mov}) and embeds a vector template matrix $\mathbb{V}$ to weight them for Equ.\ref{equ:dis2} which is equivalent with Equ.\ref{equ:dis1}. Based on this, Equ.\ref{equ:dis3} can be deduced that illustrates when the individual distance approaches its certain value, the whole will also approach the certain whole. Therefore, our COVER hypothesizes that existing a value $v$ (i.e., vector in DVF $\psi_{ab}$) can make $\sum_{i=0}^{I}\mathbb{V}^{i}\alpha^{i}=v$, the Equ.\ref{equ:dis4} will be derived from Equ.\ref{equ:dis3}, formulating distance modeling in vector regression. Compared with the binary CL, our vector CL makes a tighter generalization bound \cite{mohri2018foundations}. We have formulated and analyzed the theory foundations in the \emph{supplementary}).

\emph{Equivalency of vector regression and distance modeling.} While vector regression is not strictly equivalent to conventional distance modeling, our approach of capturing overall distance distribution enhances representation generalizability. Although Equ.\ref{equ:dis3} shows that the aggregated weighted error tends to zero, it does not guarantee that each individual error in Equ.\ref{equ:dis2} vanishes. Fortunately, since the ground truth distance $\alpha^{i}$ is task-dependent \cite{qu2005new}, our approach captures the overall distance distribution rather than enforcing exact matches for each feature pair. This flexibility allows the model to accommodate varying feature biases across different tasks, promoting a more adaptable representation.

\subsection{Self-vector regression \textbf{\emph{for extendable self-learning}}}\label{subsec:VL}
Our SeVR (Fig.\ref{fig:3}-a) constructs and regresses the ground truth vector $v$ in a self-space transformation mechanism. Given a set of medical images $\mathcal{D}$, an image $x$ is randomly sampled and cropped $x\sim \mathcal{D}$, and two transformation operations for appearance $t\sim\mathcal{T}_{ap}$ and space $\psi_{ab}\sim\mathcal{T}_{sp}$ are sampled for appearance and space transformation functions $\mathcal{T}_{ap},\mathcal{T}_{sp}$. Especially, the space transformation operation is a DVF $\psi_{ab}$ which transforms the space of the image $x$ and will play as the ground truth vectors $\{v^{i}\}_{i\in\Omega}$ in the field. Our COVER produces the transformation operations for the image $x$ and generates two views $x_a=t(x),x_b=\psi_{ab}(x)$ with different space and appearance features. Two share-weighted networks $\mathcal{N}_{\theta}$ extract the pixel-wise features with $L$ scale levels $F_a=\{f^l_a\}^{L}_{l=0}=\mathcal{N}_{\theta}(x_a),F_b=\{f^l_b\}^{L}_{l=0}=\mathcal{N}_{\theta}(x_b)$ from these views for multi-scale representation. A function $\mathcal{V}(\cdot,\cdot)$ (VPA, Sec.\ref{subsec:vpa}) formulates the relationship between the distances of these features $F_a,F_b$, predicting the DVF $\psi'_{ab}$ which indicates their correspondences. 

Our COVER learns to regress the vector indicates the correspondence to drive the distance modeling, so a vector loss $\mathcal{L}_{vec}$ calculates a mean absolute error between the predicted $\psi'_{ab}$ and the ground truth DVF $\psi_{ab}$, i.e., 
\begin{equation}\label{equ:vecloss}
  \begin{matrix}\mathcal{L}_{vec}(\psi_{ab},\psi'_{ab},\epsilon_{ab})=\sum_{i\in\{\epsilon_{ab}=1\}}|\psi^{i}_{ab}-\psi'^{i}_{ab}|\end{matrix},
\end{equation}
where the $\epsilon_{ab}$ is a mask that eliminates the content mismatch of two views $x_a,x_b$ caused by the spatial transformation and is generated from the ground truth DVF $\psi_{ab}$. It enables the network to model the correspondence degrees between one feature and a set of features (illustrated in Sec.\ref{subsec:mov}) for position correspondence, i.e., vectors. An additional consistency loss $\mathcal{L}_{con}$ is used to maintain the invariance of semantics in space transformation. It calculates a cosine similarity between the dense features from two views, i.e., $\mathcal{L}_{con}(f^4_{ab},f^4_b,\epsilon_{ab})=-\sum_{i\in\{\epsilon_{ab}=1\}}\frac{f^{4,i}_{ab}\cdot f^{4,i}_{b}}{||f^{4,i}_{ab}|| ||f^{4,i}_{b}||}$, where the $f^4_{ab}=\psi_{ab}(f^4_{a})$ is the transformed 4th level dense features $f^4_{a}$ for a consistent space of dense feature $f^4_{b}$. 

The whole optimization of our COVER iteratively minimizes the $\arg\min_{\theta}\mathcal{L}_{COVER}=\mathcal{L}_{con}+\mathcal{L}_{vec}$ for weights $\theta$. If it utilizes the SGD, its dynamics can be formulated as $\theta^{*}\leftarrow\theta-\eta\frac{\partial\mathcal{L}_{COVER}(\theta,\mathcal{D},\mathcal{T}_{sp},\mathcal{T}_{ap})}{\partial\theta}$, where the $\eta$ is the learning rate. More details are illustrated in \emph{Supplementary}.

\textbf{Discussion of properties} Our SeVR constructs a self-space transformation mechanism, enabling an \textit{extendable} self-supervised learning paradigm. It generates DVFs from random transformation as the ground truth vectors $v$ to drive the vector regression without annotation or externally paired data \cite{He_2023_CVPR,he2025homeomorphismpriorfalsepositive}. This property makes it extendable to a wide variety of medical images for large-scale training simplifying data preparation and enhancing generality. 

\subsection{Mixture of vectors \textbf{\emph{with consistent optimization flow}}}\label{subsec:mov}
As a part of the function $\mathcal{V}$, our mixture of vectors (MoV, Fig.\ref{fig:3}-b) $\mathcal{M}$ maps distances as vectors with consistent optimization flow. It constructs a vector embedding unit (VEU, b-i) and a multi-vector integration (MVI, b-ii) to encode the \emph{spatial continuity} and \emph{correspondence ambiguity}:

\textbf{Vector embedding unit (VEU) $\mathcal{U}(\cdot,\cdot)$} For a receptive field with $N\times N$ size\footnote{For a clearer illustration, we take 2D situation as example.} on position $i$ of image grid, the distances between the center feature $f^{i}_{a}$ (from view $x_a$) and the set of features $f^{N\times N}_{b}=\{f^{(n,n)}_{b}\}_{n=0}^{N}$ in the receptive field (from view $x_b$) are calculated. Inspired by the attention in transformer \cite{vaswani2017attention}, it uses the scaled dot-product attention to model their similarity to (0,1) thus achieving a distance map $D^{N\times N}$. To map the feature distances as vectors, we design a fixed vector template matrix $\mathbb{V}^{N\times N}=\{\mathbb{V}^{(n,n)}\}_{n=0}^{N}$ that indicates the vectors in which the center coordinate pointing to the coordinates in the receptive field. This vector template $\mathbb{V}^{N\times N}$ is matrix multiplied with the distance map $D^{N\times N}$ for a vector $v'^{i}_{ab}$. The function of our VEU $\mathcal{U}(\cdot,\cdot)$ is 
\begin{equation}\label{equ.veu}
  \begin{matrix}v'^{i}_{ab}=\mathcal{U}(f^{i}_{a},f^{N\times N}_{b})=\text{softmax}(\frac{f^{i}_{a}f^{\top N\times N}_{b}}{\tau})\mathbb{V}^{N\times N}\end{matrix},
\end{equation}
where the $\tau$ is a scaling factor ($\tau=\sqrt{c_{k}}$ in experiment following \cite{vaswani2017attention}). The predicted vector $v'^{i}_{ab}$ is the displacement of feature $f^{i}_{a}$ to align the most similar features in $f^{N\times N}_{b}$.

\textbf{Multi-vector integration (MVI)} Our MVI predicts and integrates multiple vectors to adapt the ambiguous correspondence in a region with continuous semantics caused by varied feature concerns. The features $f_{a},f_{b}$ with $C$ channels are divided into $J$ groups, and each group is produced by our VEU $\mathcal{U}$ for a potential vector $v'^{j}=\mathcal{U}(f_{a}[j],f_{b}[j])$ where $j$ is the index of group, thus achieving a vector group $\textbf{\text{v}}'=\{v'^{j}\}_{j}^{J}$ representing the correspondences in different feature concerns. Finally, it averages the vectors in the group for a displacement vector $u'=\frac{1}{J}\sum_{j=0}^{J}v'^{j}$ that is compatible with the ambiguity of correspondence. Calculating the displacement vectors for all positions on the image grid, our MoV will output the displacement vector field (DVF) $\psi'_{ab}=\{u'^{i}_{ab}\}_{i\in\Omega}$ that indicates the correspondence of dense features between view $x_a$ and $x_b$.

\textbf{Discussion of properties} Our MoV constructs a consistent optimization flow from vector regression to distance modeling to quantify the dispersion. It encodes the properties of \textit{spatial continuity} and \textit{correspondence ambiguity} of the pixel-wise features, bringing two advantages: \emph{1) Feature correlation preservation:} Our VEU constructs a vector template $\mathbb{V}$ that describes this continuous spatial relationship. This template $\mathbb{V}$ maps the feature distances to vectors that avoid the artificial division \cite{li2021dense,wang2022densecl,xie2021propagate} thus preserving the feature correlation. \emph{2) Bias adaptability.} Pixel-wise correspondence is uncertain and task-dependent, our MVI models this ambiguity to represent diverse feature concerns, enhancing bias adaptability in general applications.

\begin{table*}
\centering
\resizebox{\textwidth}{!}
{
\begin{tabular}{cl|cc|cc||c|ccc|c}
&
&\multicolumn{4}{c||}{\textbf{a) 2D evaluation pretrained on \cite{wang2017chestx}}}
&\multicolumn{4}{c|}{\textbf{b) 3D evaluation pretrained on \cite{marek2011parkinson}}}
&\textbf{AVG}
\\
Type
&Methods
&\textbf{SCR}$^{25\%}_{S}$
&\textbf{PDCXR}$_{C}$
&\textbf{KiPA22}$^{2D}_{S}$
&\textbf{FIVES}$_{S}$
&\textbf{CANDI}$_{S}$
&\textbf{FeTA21}$_{S}$
&\textbf{KiPA22}$^{3D}_{S}$
&\textbf{STOIC}$_{C}$
&\emph{Score \%}
\\
\hline
\color{gray}-
&\color{gray}Scratch
&\color{gray}81.8
&\color{gray}90.4
&\color{gray}74.1
&\color{gray}79.4
&\color{gray}84.0
&\color{gray}56.9
&\color{gray}72.4
&\color{gray}72.0
&\color{gray}76.4
\\
\cdashline{1-11}[0.8pt/2pt]
GL
&Denosing \cite{vincent2010stacked}
&83.9$\color{blue}_{(+1.9)}$
&92.0$\color{blue}_{(+1.6)}$
&60.3$_{(-13.8)}$
&77.8$_{(-1.6)}$
&83.7$_{(-0.3)}$
&52.9$_{(-4.0)}$
&70.0$_{(-2.4)}$
&65.9$_{(-6.1)}$
&73.3$_{(-3.1)}$
\\
&In-painting \cite{pathak2016context}
&85.1$\color{blue}_{(+3.3)}$
&93.9$\color{blue}_{(+3.5)}$
&64.4$_{(-9.7)}$
&78.9$_{(-0.5)}$
&88.5$\color{blue}_{(+4.5)}$
&54.4$_{(-2.5)}$
&69.7$_{(-2.7)}$
&67.2$_{(-4.8)}$
&75.3$_{(-1.1)}$
\\
&Models Genesis \cite{zhou2019models}
&86.1$\color{blue}_{(+4.3)}$
&92.6$\color{blue}_{(+2.2)}$
&66.6$_{(-7.5)}$
&79.6$\color{blue}_{(+0.2)}$
&88.7$\color{blue}_{(+4.7)}$
&55.8$_{(-1.1)}$
&75.8$\color{blue}_{(+3.4)}$
&75.3$\color{blue}_{(+3.3)}$
&77.6$\color{blue}_{(+1.2)}$
\\
&Rotation \cite{komodakis2018unsupervised}
&80.5$_{(-1.3)}$
&89.9$_{(-0.5)}$
&69.7$_{(-4.4)}$
&80.3$\color{blue}_{(+0.9)}$
&89.4$\color{blue}_{(+5.4)}$
&58.7$\color{blue}_{(+1.8)}$
&77.4$\color{blue}_{(+5.0)}$
&68.8$_{(-3.2)}$
&76.8$\color{blue}_{(+0.4)}$
\\
\cdashline{1-11}[0.8pt/2pt]
BCL
&SimSiam \cite{Chen2021CVPR}
&87.2$\color{blue}_{(+5.4)}$
&92.2$\color{blue}_{(+1.8)}$
&72.6$_{(-1.5)}$
&84.3$\color{blue}_{(+4.9)}$
&87.3$\color{blue}_{(+3.3)}$
&58.7$\color{blue}_{(+1.8)}$
&83.8$\color{blue}_{(+11.4)}$
&69.5$_{(-2.5)}$
&79.5$\color{blue}_{(+3.1)}$
\\
&BYOL \cite{grill2020bootstrap}
&89.4$\color{blue}_{(+7.6)}$
&86.6$_{(-3.8)}$
&74.1$_{(0)}$
&83.3$\color{blue}_{(+3.9)}$
&89.7$\color{blue}_{(+5.7)}$
&59.2$\color{blue}_{(+2.3)}$
&83.6$\color{blue}_{(+11.2)}$
&74.0$\color{blue}_{(+2.0)}$
&80.0$\color{blue}_{(+5.4)}$
\\
&SimCLR \cite{chen2020simple}
&89.0$\color{blue}_{(+7.2)}$
&\cellcolor[HTML]{D3E6F1}94.7$\color{blue}_{(+4.3)}$
&74.4$\color{blue}_{(+0.3)}$
&84.5$\color{blue}_{(+5.1)}$
&89.2$\color{blue}_{(+5.2)}$
&53.4$_{(-3.5)}$
&78.9$\color{blue}_{(+6.5)}$
&60.7$_{(-11.3)}$
&78.1$\color{blue}_{(+1.7)}$
\\
&MoCov2 \cite{chen2020improved}
&84.3$\color{blue}_{(+2.5)}$
&93.2$\color{blue}_{(+2.8)}$
&69.6$_{(-4.5)}$
&80.7$\color{blue}_{(+1.3)}$
&89.7$\color{blue}_{(+5.7)}$
&61.5$\color{blue}_{(+4.6)}$
&78.0$\color{blue}_{(+5.6)}$
&74.8$\color{blue}_{(+2.8)}$
&79.0$\color{blue}_{(+2.6)}$
\\
&DeepCluster \cite{caron2018deep}
&84.0$\color{blue}_{(+2.2)}$
&93.3$\color{blue}_{(+2.9)}$
&72.7$_{(-1.4)}$
&81.6$\color{blue}_{(+2.2)}$
&89.8$\color{blue}_{(+5.8)}$
&57.4$\color{blue}_{(+0.5)}$
&79.7$\color{blue}_{(+7.3)}$
&65.6$_{(-6.4)}$
&78.0$\color{blue}_{(+1.6)}$
\\
\cdashline{1-11}[0.8pt/2pt]
DBCL
&VADeR \cite{o2020unsupervised}
&85.2$\color{blue}_{(+3.4)}$
&92.8$\color{blue}_{(+2.4)}$
&62.8$_{(-11.3)}$
&78.9$_{(-0.5)}$
&87.4$\color{blue}_{(+3.4)}$
&43.1$_{(-13.8)}$
&72.1$_{(-0.3)}$
&73.2$\color{blue}_{(+1.2)}$
&74.4$_{(-2.0)}$
\\
&DenseCL \cite{wang2022densecl}
&85.0$\color{blue}_{(+3.2)}$
&92.4$\color{blue}_{(+2.0)}$
&70.8$_{(-3.3)}$
&79.2$_{(-0.2)}$
&87.7$\color{blue}_{(+3.7)}$
&43.7$_{(-13.2)}$
&74.0$\color{blue}_{(+1.6)}$
&58.8$_{(-13.2)}$
&74.0$_{(-2.4)}$
\\
&SetSim \cite{wang2022exploring}
&85.2$\color{blue}_{(+3.4)}$
&93.9$\color{blue}_{(+3.5)}$
&70.8$_{(-3.3)}$
&80.1$\color{blue}_{(+0.7)}$
&88.4$\color{blue}_{(+4.4)}$
&58.7$\color{blue}_{(+1.8)}$
&73.5$\color{blue}_{(+1.1)}$
&60.1$_{(-11.9)}$
&76.3$_{(-0.1)}$
\\
&DSC-PM \cite{li2021dense}
&90.5$\color{blue}_{(+8.7)}$
&91.8$\color{blue}_{(+1.4)}$
&77.2$\color{blue}_{(+3.1)}$
&83.8$\color{blue}_{(+4.4)}$
&88.5$\color{blue}_{(+4.5)}$
&52.2$_{(-4.7)}$
&79.0$\color{blue}_{(+6.6)}$
&59.5$_{(-12.5)}$
&77.8$\color{blue}_{(+1.2)}$
\\
&PixPro \cite{xie2021propagate}
&91.5$\color{blue}_{(+9.7)}$
&93.0$\color{blue}_{(+2.6)}$
&73.6$_{(-0.5)}$
&84.3$\color{blue}_{(+4.9)}$
&89.9$\color{blue}_{(+5.9)}$
&60.7$\color{blue}_{(+3.8)}$
&80.0$\color{blue}_{(+7.6)}$
&75.1$\color{blue}_{(+3.1)}$
&81.0$\color{blue}_{(+4.6)}$
\\
&Chaitanya et al. \cite{chaitanya2020contrastive}
&87.3$\color{blue}_{(+5.5)}$
&90.0$_{(-0.4)}$
&76.5$\color{blue}_{(+2.4)}$
&84.9$\color{blue}_{(+5.5)}$
&87.4$\color{blue}_{(+3.4)}$
&53.4$_{(-3.5)}$
&70.7$_{(-1.7)}$
&67.8$_{(-4.2)}$
&77.3$\color{blue}_{(+0.9)}$
\\
\cdashline{1-11}[0.8pt/2pt]
VR
&GVSL \cite{He_2023_CVPR}
&89.7$\color{blue}_{(+7.9)}$
&92.5$\color{blue}_{(+2.1)}$
&78.9$\color{blue}_{(+4.8)}$
&\cellcolor[HTML]{D3E6F1}86.2$\color{blue}_{(+6.8)}$
&89.1$\color{blue}_{(+5.1)}$
&\cellcolor[HTML]{D3E6F1}62.6$\color{blue}_{(+5.7)}$
&84.3$\color{blue}_{(+11.9)}$
&75.4$\color{blue}_{(+3.4)}$
&82.3$\color{blue}_{(+5.9)}$
\\
&GEMINI \cite{he2025homeomorphismpriorfalsepositive}
&\cellcolor[HTML]{D3E6F1}92.4$\color{blue}_{(+10.6)}$
&92.9$\color{blue}_{(+2.5)}$
&\cellcolor[HTML]{D3E6F1}79.1$\color{blue}_{(+5.0)}$
&85.3$\color{blue}_{(+5.9)}$
&\cellcolor[HTML]{FFCCC9}\textbf{90.0$\color{blue}_{(+6.0)}$}
&61.7$\color{blue}_{(+4.8)}$
&\cellcolor[HTML]{D3E6F1}85.0$\color{blue}_{(+12.6)}$
&\cellcolor[HTML]{D3E6F1}79.5$\color{blue}_{(+7.5)}$
&\cellcolor[HTML]{D3E6F1}83.2$\color{blue}_{(+6.8)}$
\\
\cdashline{1-11}[0.8pt/2pt]
\textit{VCL}
& \textbf{COVER (Ours)}
&\cellcolor[HTML]{FFCCC9}\textbf{94.0$\color{blue}_{(+12.2)}$}
&\cellcolor[HTML]{FFCCC9}\textbf{95.9$\color{blue}_{(+5.5)}$}
&\cellcolor[HTML]{FFCCC9}\textbf{80.0$\color{blue}_{(+5.9)}$}
&\cellcolor[HTML]{FFCCC9}\textbf{87.2$\color{blue}_{(+7.8)}$}
&\cellcolor[HTML]{D3E6F1}89.9$\color{blue}_{(+5.9)}$
&\cellcolor[HTML]{FFCCC9}\textbf{63.6$\color{blue}_{(6.7)}$}
&\cellcolor[HTML]{FFCCC9}\textbf{85.2$\color{blue}_{(+12.8)}$}
&\cellcolor[HTML]{FFCCC9}\textbf{80.4$\color{blue}_{(+8.4)}$}
&\cellcolor[HTML]{FFCCC9}\textbf{84.5$\color{blue}_{(+8.1)}$}
\end{tabular}
}
\caption{\textbf{Comparison study:} Our COVER achieves the best performance compared with 17 methods on 8 downstream tasks across a) 2D and b) 3D, showing our great adaptation ability. ``AVG" is the average rank of the row. The blue values are the improvement compared with the ``Scratch" and the bold values are the highest score. ``S" and ``C" are the segmentation and classification tasks.}
\label{tab:metrics1}
\end{table*}
\subsection{Vector pyramid aggregation \textbf{\emph{adapts granularity}}}\label{subsec:vpa}
Stacking our MoV $\mathcal{M}$ in a pyramid-like architecture, our vector pyramid aggregation (VPA, Fig.\ref{fig:3}-c) formulates the final mapping function $\mathcal{V}$ to model the multiscalarity of correspondence efficiently, training a granularity adaptability. This architecture divides multi-level correspondences according to the scale levels of the dense features. By fusing these scale levels, the VPA will output the DVF $\psi'_{ab}$ that models both global and local correspondences. In 0th level, dense features $f^{0}_a,f^{0}_b$ are put into our MoV module for a 0th-level DVF $\psi'^{0}_{ab}=\mathcal{M}(f^0_a,f^0_b)$ with a global correspondence for high-level features. Then, in 1st level, the $\psi'^0_{ab}$ transforms the 1st-level dense features $f^1_a$ to bridge the space gap between $f^1_a$ and $f^1_b$ on the 0th level. The features are further produced by our MoV and further fused with the $psi'^0_{ab}$ for the 1st-level DVF $\psi'^{1}_{ab}=\mathcal{M}(\psi'^{0}_{ab}(f^{1}_{a}),f^{1}_{b})\bigodot\psi'^{0}_{ab}$, where the ``$\bigodot$" is the fusion operation illustrated in \emph{Supplementary}. The remaining levels are calculated similarly so that the final DVF $\psi'_{ab}$ integrated multiscale vectors are output after a chain calculation, i.e. 
\begin{align}\label{equ:vpa}
  &\begin{matrix}\psi'_{ab}=\mathcal{V}(F_a,F_b)=H(\{\psi'^{l}_{ab}\}^{L}_{l=0}),\ \text{where}\end{matrix}\\
  &\begin{matrix}\psi'^{0}_{ab}=\mathcal{M}(f^0_a,f^0_b) \end{matrix}\notag \\
  &  \begin{matrix}\psi'^{l}_{ab}=\mathcal{M}(\psi'^{l-1}_{ab}(f^{l}_{a}),f^{l}_{b})\bigodot\psi'^{l-1}_{ab},\ l=1,2,...,L-1,\end{matrix}\notag
\end{align}
where the $H$ is the chain calculation across all $L$ levels to output the final DVF $\psi'_{ab}$. Linear upsampling and value doubling are applied to bridge the scale gap of the DVFs.

\textbf{Discussion of properties} Correspondence occurs at multiple scales for different semantic granularity, so our VPA formulates this \emph{multiscalarity} in a pyramid-like architecture, bringing the advantages: \textit{1) Low computational cost:} It enables the mapping function $\mathcal{V}$ in a small receptive field at each level for a large whole receptive field reducing the computation. \textit{2) Multi-scale representation:} It learns the multi-scale features with multiple semantic granularities improving the granularity adaptability.

\section{Experiments}
\subsection{Comparison study}
\textbf{4.1.1 Experiment protocol} This study has made sufficient experiments to ensure a complete evaluation. Following is the overview protocol and details are in our \emph{Supplementary}.

\begin{figure}
    \centering
    \includegraphics[width=1\linewidth]{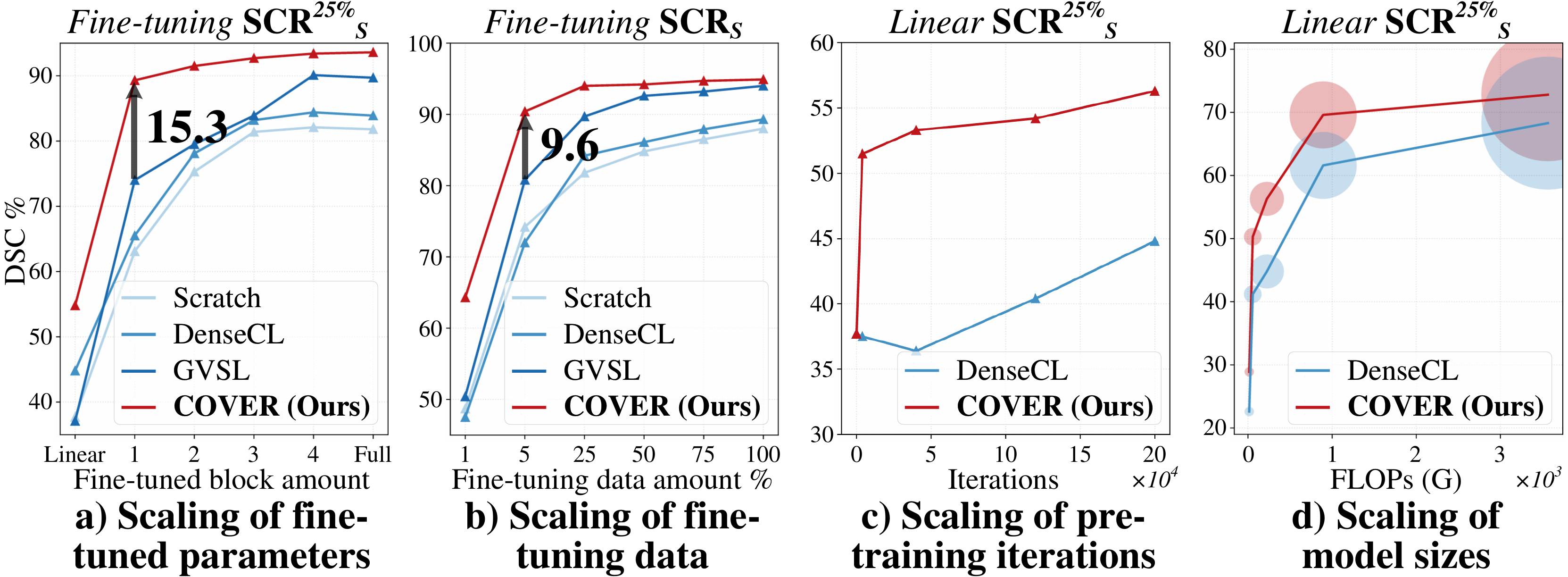}
    \caption{\textbf{Implementation analysis:} Our COVER has great properties on the scaling of a) fine-tuned parameters, b) fine-tuning data, c) pretraining iterations, and d) model sizes.}
    \label{fig:7}
\end{figure}

\emph{\textbf{1) Materials:}} This study evaluated our COVER across 8 tasks across 2D and 3D evaluations with 4 modalities to showcase its versatility and advantages. \textbf{a)} \emph{2D evaluation:} It pretrained networks on the ChestX-ray6 dataset \cite{wang2017chestx} containing 112,120 chest X-ray images. These pretrained models are fine-tuned on four tasks (SCR$^{25\%}_{S}$ \cite{van2006segmentation} (X-ray), PDCXR$_{C}$ \cite{kermany2018identifying} (X-ray), FIVES$_{S}$ \cite{jin2022fives} (fundus), KiPA22$^{2D}_{S}$ \cite{he2021meta} (CT)). \textbf{b)} \emph{3D evaluation:} It pretrained networks on 837 3D T1 brain MR images from the PPMI database \cite{marek2011parkinson}. These pretrained models are then fine-tuned to three datasets with four tasks (CANDI$_{S}$ \cite{kennedy2012candishare} (MR), FeTA21$_{S}$ \cite{payette2023fetal} (MR), STOIC$_{C}$ \cite{revel2021study} (CT), KiPA22$^{3D}_{S}$ \cite{he2021meta} (CT)). The evaluations involve segmentation (S) and classification (C) tasks. To evaluate the data amount robustness, it used 25\% of the training set in SCR$_{S}$, and more data amount evaluations are performed in our analysis (Fig.\ref{fig:7} b)).

\emph{\textbf{2) Comparisons:}} This study benchmarked COVER with 17 recent or classic methods across four categories: GL \cite{vincent2010stacked,pathak2016context,zhou2019models,komodakis2018unsupervised}, binary CL (BCL) \cite{Chen2021CVPR,grill2020bootstrap,chen2020simple,chen2020improved,caron2018deep}, dense binary CL (DBCL) \cite{o2020unsupervised,wang2022densecl,wang2022exploring,xie2021propagate,li2021dense,chaitanya2020contrastive}, and vector regression (VR) \cite{He_2023_CVPR,he2025homeomorphismpriorfalsepositive}. It used 2D and 3D U-Nets \cite{ronneberger2015u} as the backbone $\mathcal{N}_{\theta}$ (for the methods with global prediction, like MoCov2, it adopted the encoder part). For a fair comparison, full-parameter fine-tuning is employed in downstream tasks. 

\emph{\textbf{3) Implementation and evaluation metrics:}} Following \cite{He_2023_CVPR}, all tasks were implemented by PyTorch and optimized by Adam with a learning rate of 10$^{-4}$. The models were pretrained with $2\times10^{5}$ iterations for pretraining tasks and fine-tuned with $4\times10^{4}$ iterations for downstream tasks. Training and testing were implemented on NVIDIA A100 SXUM4 GPU with 40 GB memory. This study used the Dice coefficient (DSC) for segmentation tasks and area under curve (AUC) for classification tasks \cite{taha2015metrics}.

\textbf{4.1.2 Quantitative analysis} Our great adaptability for the downstream tasks shown in Tab. \ref{tab:metrics1} demonstrates our COVER's powerful general representation for medical vision. Two observations can be discovered: 

\textbf{\emph{1) Cross-scale transferability}}. The SOTA performances on small and large targets show our powerful cross-scale transferability owing to the learned multiscalarity from our VPA \textbf{a.} For small structures like the vessels (FIVES$_{S}$) and brain tissues (CANDI$_{S}$ and FeTA$_{S}$), our approach achieved 87.2\%, 89.9\%, and 63.6\% DSC with the improvements of 7.8\%, 5.9\%, and 6.7\% over the ``Scratch". These gains stem from learning fine-grained representations at lower-scale levels. \textbf{b.} For large structures including the chest regions (SCR$^{25\%}_{S}$) and kidney structures (KiPA22$^{2D}_{S}$, and KiPA22$^{3D}_{S}$), our model achieved 94.0\%, 80.0\%, and 85.2\% DSC improving by over 5\% compared to ``Scratch". This is attributed to leveraging high-scale levels for coarse-grained representation. \textbf{c.} For classification tasks (PDCXR$_{C}$ and STOIC$_{C}$), our COVER achieved AUC improvements of 5.5\% and 8.4\% by learning global representations at level 0. Although the GVSL and GEMINI also have improvements across scales, their inconsistent optimization flow limits the representation potential of pretraining data, resulting in an average score over 1\% lower than COVER. The BCL methods struggle to enhance all tasks, for example, the DenseCL improved 3.2\% DSC on SCR, but it reduced 3.3\% DSC on KiPA22$^{2D}_{S}$. This arises from their over-dispersion, which introduces biases and compromises on incompatible tasks.

\textbf{\emph{2) Cross-scene adaptability}}. The significant improvements on consistent (SCR$^{25\%}_{S}$, PDCXR$_{C}$, CANDI$_{S}$) and inconsistent (KiPA22$^{2D}_{S}$, FIVES$_{S}$, FeTA$_{S}$, KiPA22$^{3D}_{S}$, STOIC$_{C}$) scenes relative to the pretraining data illustrates our strong cross-scene adaptability. Across all tasks in both 2D and 3D evaluations, COVER outperformed the ``Scratch" by more than 5\% and achieved the highest average scores (84.5\%). This adaptability arises from our distance modeling which quantifies the dispersion in embedding space, enabling the disentanglement of underlying explanatory factors hidden in low-level sensory data \cite{bengio2013representation}. These learned factors capture the intrinsic properties of medical images, enhancing versatility across diverse scenes. Although most methods improve performance in consistent scenes by leveraging learned transformable knowledge, they struggle in inconsistent scenes. Notably, BCL and DBCL even degrade performance in some tasks due to their over-dispersion in binary CL, which amplifies biases in pretraining data. Especially, DenseCL and VADeR perform particularly poorly in inconsistent scenes like FeTA21$_{S}$, which reduce over 10\% DSC.

\subsection{Ablation study and model analysis}
\textbf{4.3.1 Component ablation} The ablation study of the components in our COVER on the 2D evaluation of SCR$^{25\%}_{S}$ task shows the effectiveness of our proposed innovations. This study took the learning of the consistency loss $\mathcal{L}_{con}$ in pixel-wise as our base model, and gradually added our proposed components. The results are as follows:
\begin{table}[H]
  \centering
\resizebox{0.9\linewidth}{!}
{
  \begin{tabular}{c|ccccc}
    & \emph{base.} $\mathcal{L}_{con}$
    & ...+VEU (SeVR)
    & ...+VPA
    & ...+MVI
    \\
    \hline
    DSC\%
    & 91.8
    & 92.9
    & 93.4
    & 94.0
  \end{tabular}
  }
\end{table}
\noindent The base model $\mathcal{L}_{con}$ has 91.8\% DSC that is higher than the BYOL's 89.4\% owing to its pixel-wise representation. When adding our VEU, it effectively improves the discrimination, achieving a 1.1\% improvement. Our VPA enhances the multi-scale representation so that the model further improves 0.5\% DSC. Finally, our MVI models the ambiguity of the correspondence for potentially varied feature concerns, thus achieving the highest 94.0\% DSC.

\textbf{4.3.2 Hyper-parameter ablation} The receptive field $N$ in VEU, the amount of VEUs $J$ in MVI are two new hyper-parameters in our COVER. This study took an ablation of them on the SCR$^{25\%}_{S}$ via \emph{linear evaluation} to demonstrate their properties. The results are as follows:
\begin{table}[H]
  \centering
\resizebox{1\linewidth}{!}
{
  \begin{tabular}{c|ccccc}
    $N=$
    & $3\times3$
    & $5\times5$
    & $7\times7$
    & $9\times9$
    \\
    DSC\%
    & 44.9
    & 48.4
    & 54.8
    & 35.1
    \\
    \hline
    $J=$
    & $[2,2,2,1,1]$
    & $[4,4,4,1,1]$
    & $[8,8,8,2,2]$
    & $[12,12,12,3,3]$
    \\
    DSC\%
    & 55.3
    & 56.3
    & 54.8
    & 48.6
  \end{tabular}
  }
\end{table}
\noindent For the receptive field size $N$, when enlarging it, its performance increases to $7\times7$ and then decreases when it is $9\times9$. Because the increase of $N$ will adapt to large spatial correspondence, improving the pretrained representation. However, too large $N$ will introduce more ambiguous semantics, such as vascular with similar features, which will mislead the learning of correspondence. For the amount of VEUs $k$, it also increases and then decreases. When it is [4,4,4,1,1] for the scale levels, the model achieves the best ability. Too many vectors will smooth the optimization, thus weakening the discrimination of learned features.
\begin{figure}
    \centering
    \includegraphics[width=1\linewidth]{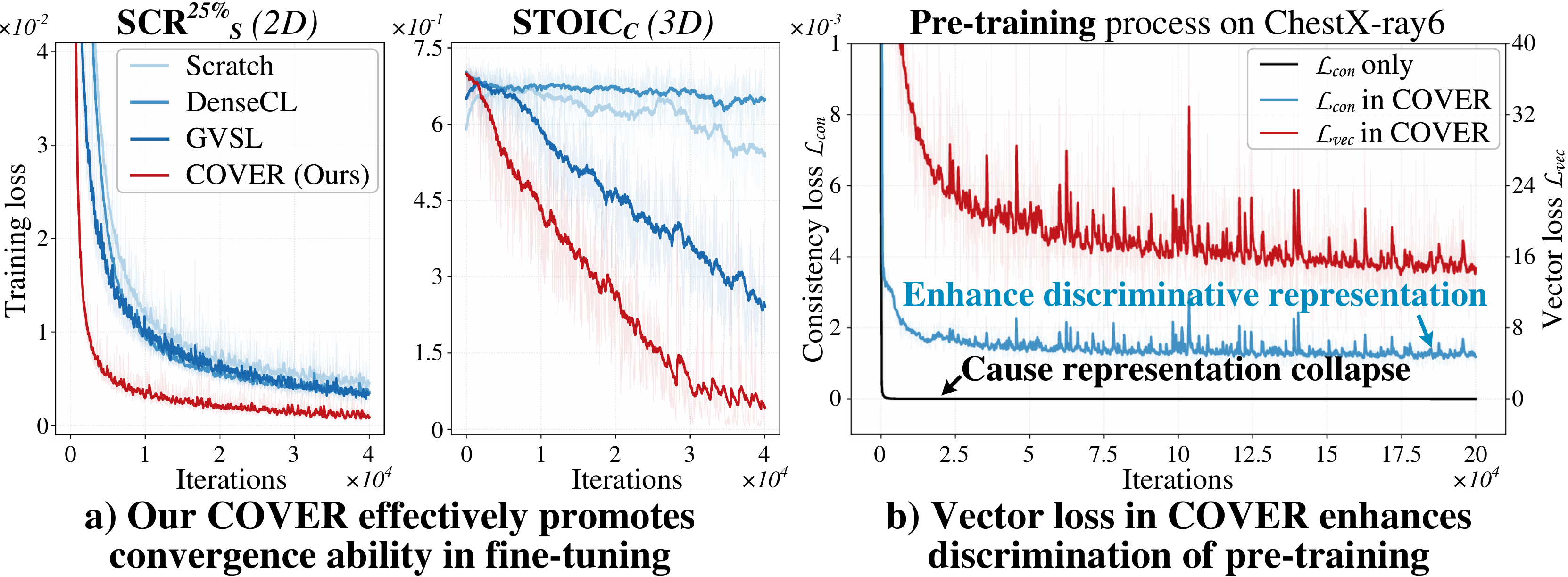}
    \caption{\textbf{Convergence analysis:} a) Our COVER is able to promote convergence ability in fine-tuning. b) In the pretraining, the consistency learning causes representation collapse, and when adding our vector regression, the discrimination is enhanced.}
    \label{fig:4}
\end{figure}

\textbf{4.3.3 Implementation analysis} This experiment analyzed four implementation factors (Fig.\ref{fig:7}): a) Fine-tuned parameter amount: Our COVER effectively reduces the fine-tuned parameters. It only fine-tunes one U-Net module to achieve a very significant performance improvement (15.3\% DSC higher than GVSL), quickly reaching the bottleneck level that SCR$^{25\%}_{S}$ task can reach.  b) Fine-tuning data amount: Our COVER effectively reduces the fine-tuning data requirement. Only utilizing 5\% training data, our COVER can achieve the DSC close to 90\% that GVSL and DenseCL have to use 25\% and 100\% data. c) pretraining iteration: With the progress of the training iterations, the performance (linear evaluation) on SCR$^{25\%}_{S}$ of our COVER gradually improves and eventually tends to be flat. This illustrates that more training is beneficial for the vector CL, which makes the pretrained models capture more diverse features in detail. d) Model size: With the enlarging of model size, the larger capacity enables our COVER to capture more abundant feature information in the pretraining. This makes the model gain more powerful representability in downstream tasks only with a linear layer.

\textbf{4.3.4 Convergence analysis} This study analyzed the convergence performance of our COVER in both pretraining and fine-tuning phases (Fig.\ref{fig:4}): \emph{a) Fine-tuning:} In downstream tasks, the network pretrained by our COVER exhibits superior convergence ability. Compared with the ``Scratch", GVSL, and DenseCL, COVER achieves a significantly lower loss value more quickly in both 2D segmentation (SCR) and 3D classification (STOIC) tasks. This advantage arises from our powerful pixel-wise feature learning that enables the pretrained network to capture the detailed information internal medical images. \emph{b) pretraining:} COVER improves the discrimination of the representation. When relying solely on the consistency loss $\mathcal{L}_{con}$, the loss quickly collapses to zero, hindering subsequent knowledge learning. By integrating vector regression $\mathcal{L}_{vec}$, the model learns feature distances, stabilizing the consistency loss with fluctuations and enhancing the discrimination.
\begin{figure}
    \centering
    \includegraphics[width=1\linewidth]{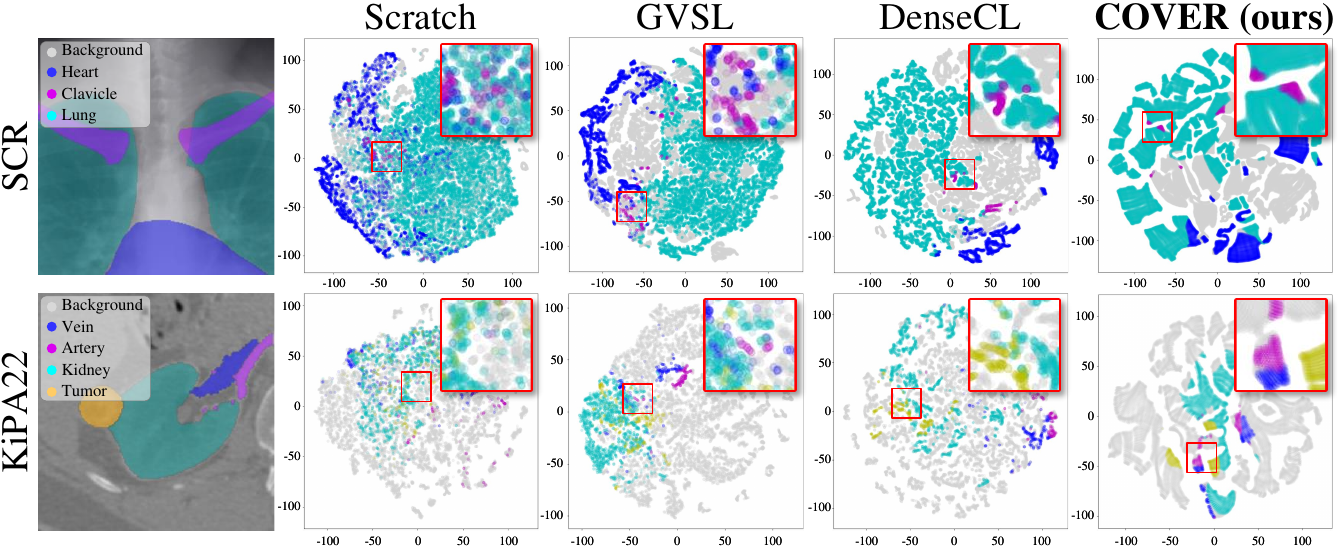}
    \caption{\textbf{Cluster analysis:} Our COVER has strong clustering capabilities by quantifying the dispersion degree of features.}
    \label{fig:5}
\end{figure}

\textbf{4.3.5 Computation efficiency} This study calculated the FLOPs to evaluate the computation efficiency with the expanding of receptive field. For a significant analysis, we focused on the computation in head part, i.e., our VPA. It compared two methods: a. VPA; b. directly taking a large vector template at level 4 (named ``Direct"). The results are:
\begin{figure}[H]
    \centering
    \includegraphics[width=1\linewidth]{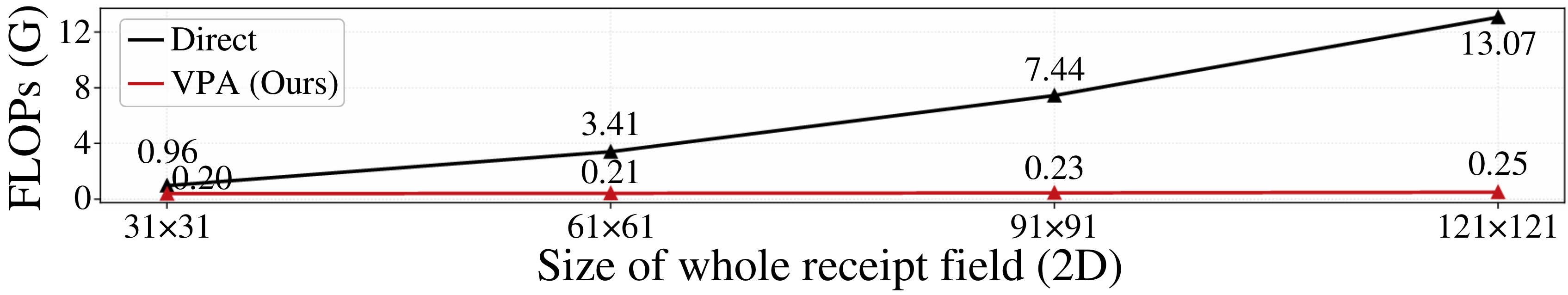}
\end{figure}
\noindent When expanding the whole receptive field, our VPA maintains a stable computational cost of 0.2, whereas the ``Direct" approach shows an exponential increase. Especially when it is $121\times121$, the ``Direct" has more than 52 times our consumption. Our VPA is able to expand the receptive field for a wider contrast with low additional consumption.

\textbf{4.3.6 Distributed clusters from COVER} The clusters show that our COVER has strong clustering capabilities by quantifying the dispersion of features. Using pretrained models from our 2D evaluation, this study extracted pixel-wise features from the SCR and KiPA22 datasets and visualized their distributions via t-SNE \cite{van2008visualizing} in Fig.\ref{fig:5}. It compared the clustering patterns of ``Scratch", GVSL, and DenseCL with our COVER, revealing two key advantages: a) Continuous feature distribution: Our COVER predicts features that are distributed continuously in the embedded space. This continuity arises from our vector CL, which models the feature distance and avoids the excessive discretization that blurs small or subtle features. However, DenseCL produces a lot of discrete clusters with small gaps between them, i.e., over-dispersion, leading to fragmented representation and reduced transferability. b) Effective aggregation: Our COVER aggregates features with the same semantics, while other methods show mixed regions. In both SCR and KiPA22 datasets, ``Scratch" and GVSL exhibit overlapped clusters, as shown in the enlarged patches. Although DenseCL performs well on SCR, it fails on KiPA22, mixing features in this inconsistent scene with the pretraining. Our COVER achieves effective aggregation across datasets because our modeling of distance mitigates the overfitting to the pretraining scene.

\begin{figure}
    \centering
    \includegraphics[width=1\linewidth]{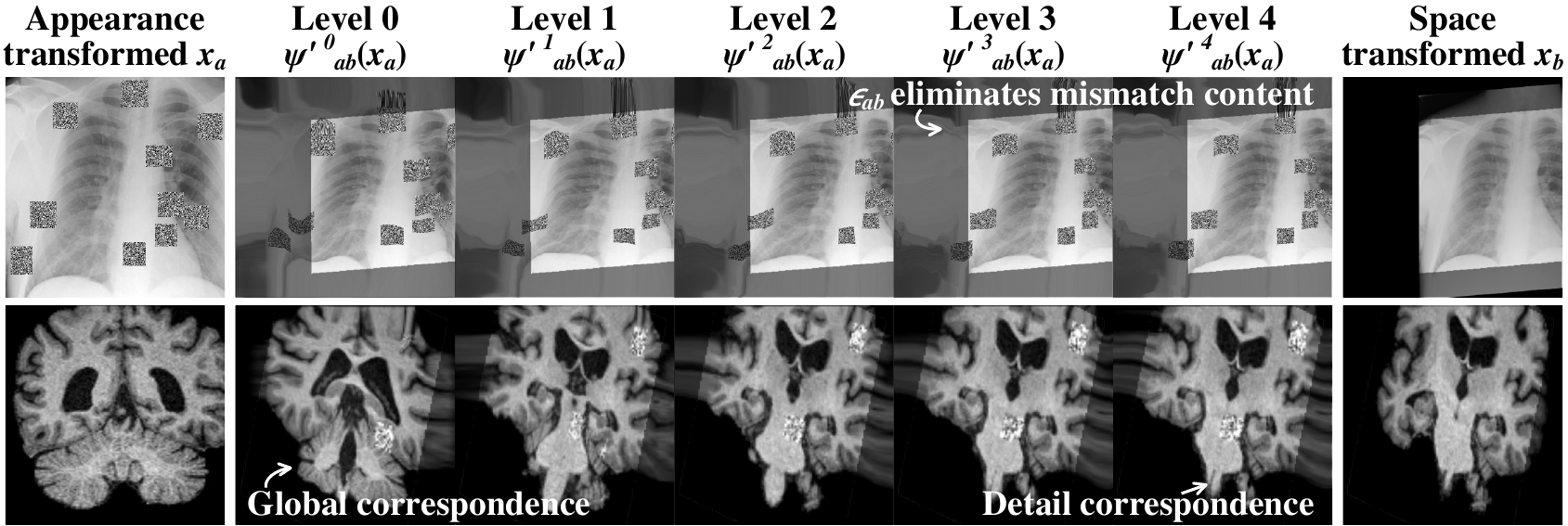}
    \caption{\textbf{Multi-scale vectors analysis:} Our VPA predicts multi-level correspondence for both global and detailed representation. }
    \label{fig:6}
\end{figure}
\textbf{4.3.7 Analysis of multi-scale vectors in our VPA} This study analyzes the vectors at each scale level in our VPA by visualizing spatial transformed images via the predicted DVFs (Fig.\ref{fig:6}) in our 2D (chest X-ray) and 3D (brain MR) evaluations. In level 0, the DVF predicted from global features is able to align the images on the whole, like the rotation, scaling, etc., driving the learning of global representation. With the expansion of the scales, the correspondences are gradually refined so the details between the images are aligned. Especially, in level 0, the images have high alignment degree on the boundary, so that the models will effectively learn the representation of small or subtle features.

\section{Conclusion}
In this paper, we reformulate the standard CL as a vector regression problem, vector CL, providing a solution for ``over-dispersion" problem caused by binary CL in pixel-wise medical visual pretraining. Our proposed \textbf{CO}ntrast in \textbf{VE}ctor \textbf{R}egression (\textbf{COVER}) framework implemented the reformulated vector CL, modeling feature distance for CL in a vector regression process. It enables quantitative dispersion of features which improves representation learning for medical vision applications. Although the unique properties of our vector CL have bright powerful performance, important future work is to involve a larger-scale medical image dataset in training, fostering the development of a large pixel-wise foundation model in the medical field. Beyond advancing MVFM \cite{he2024foundation} development, COVER opens new frontiers in learning semantically coherent representations for continuous signals.

\paragraph*{Acknowledgment} This research was supported by the National Science Foundation grants (2306545) and National Institutes of Health grants (1R01HL177813-01, 1R01HL173186-01A1).

\appendix
\vspace{1.5em}
\begin{center}
    {\Large \textbf{Supplementary Material}}
\end{center}
\section{Theoretical Foundation}\label{sec:A}
Existing pixel-wise contrastive learning (CL) still follows binary CL \cite{tian2020makes,Chen2021CVPR,chen2020simple,oquab2024dinov2,grill2020bootstrap,caron2018deep,chen2020improved} that pulls positive pairs together and push negative pairs apart, being extremely limited by the over-dispersion problem, leading to a loss of local semantic continuity—an issue especially detrimental to medical image analysis. In contrast, we introduce a theoretical framework to derive the generalization bound for pixel-wise CL. We propose a novel CL paradigm, vector CL, which models the distances between pixel-wise features through a structured mapping function, controlling dispersion while preserving local correlations.
\subsection{ Recap of the Over-Dispersion Problem and Rademacher Complexity}
\paragraph{Foundation of Contrastive Learning} Let $x\in\mathcal{X}$ denote an input medical image, and assume that a network $f:\mathcal{X}\rightarrow\mathbb{R}^d$ extracts pixel-wise features $\{ f(x)_i\}_{i=1}^{N}$ from the image, where $N$is the number of pixels. Define a similarity metric $\langle f_\theta(x_i),f_\theta(x_j)\rangle$ to measure the distance between pixel-wise features.  In binary CL, the loss $\mathcal{L}_{BCL}$ for a given pixel on position $i$ can be expressed as an InfoNCE-style \cite{oord2018representation} objective:
\begin{equation}
  \begin{matrix}\mathcal{L}_{BCL}= -\log \frac{e^{\langle f(x)_i, f(x^+)_i \rangle/\tau}}{e^{\langle f(x)_i, f(x^+)_i \rangle/\tau} +\sum_{j}e^{\langle f(x)_i, f(x^-)_j \rangle/\tau}},\end{matrix}
\end{equation}
where $x^+$ is a positive (similar) view of $x$ and $x^-$ represents negative samples; $\tau$ is a temperature parameter. For pixel-wise tasks, we additionally assume a local smoothness constraint: for any pixel on position $i$ and its local neighborhood $\mathcal{N}(i)$,
\begin{equation}
\langle f(x)_i,f(x)_j\rangle \leq \delta, \quad \forall j \in \mathcal{N}(i),
\end{equation}
with $\delta$ being a small constant relative to the range of possible feature values $\Delta$. The upper bound of the $\delta$ to describe the dispersion of the local features: for any two pixels on position $j,k$ in the local neighborhood $\mathcal{N}(i)$,
\begin{equation}
\delta=\max_{j,k\in\mathcal{N}(i)}|\langle f(x)_i,f(x)_j\rangle-\langle f(x)_i,f(x)_k\rangle|.
\end{equation}

\paragraph{Foundation of Rademacher Complexity} Let $\mathcal{F}$ denote the hypothesis space of feature extractors. The empirical Rademacher complexity \cite{mohri2018foundations} is defined as
\begin{equation}
  \mathfrak{R}_n(\mathcal{F}) = \mathbb{E}_{\sigma}\left[\sup_{f \in \mathcal{F}} \frac{1}{n} \sum_{i=1}^{n} \sigma_i f(x_i)\right],
\end{equation}
where $\{\sigma_i\}$ are independent Rademacher variables. For pixel-wise learning, one might consider $nN$ samples (with $n$ images and $N$ pixels per image). However, the local smoothness constraint effectively reduces the ``degrees of freedom" in each local neighborhood \cite{bartlett2002rademacher,koltchinskii2006local}. We can thus define an effective local Rademacher complexity as
\begin{equation}
\mathfrak{R}_{n,\text{local}}(\mathcal{F}) \leq \left(\frac{\delta}{\Delta}\right) \mathfrak{R}_n(\mathcal{F}),
\end{equation}
where the factor $\delta/\Delta$ represents the reduction in complexity due to local smoothness.

By standard generalization theory \cite{mohri2018foundations}, for any $f \in \mathcal{F}$, with high probability (at least $1-\varepsilon$), the generalization error satisfies
\begin{equation}
R(f) \leq \hat{R}(f) + 2\mathfrak{R}_{n,\text{local}}(\mathcal{F}) + O\left(\sqrt{\frac{\log(1/\varepsilon)}{nN}}\right),
\end{equation}
or equivalently,
\begin{equation}\label{equ:gen}
R(f) \leq \hat{R}(f) + 2\left(\frac{\delta}{\Delta}\right) \mathfrak{R}_n(\mathcal{F}) + O\left(\sqrt{\frac{\log(1/\varepsilon)}{nN}}\right).
\end{equation}
Thus, reducing the maximum local difference $\delta$ through better modeling of local continuity directly tightens the generalization bound.

\paragraph{Over-Dispersion Problem} In the binary CL, since there is no explicit control over how close or dispersed feature representations are within a local neighborhood $\mathcal{N}(i)$, the difference between adjacent pixel features $\delta_{BCL}$ is unrestricted and can increase without bound. Therefore, the features will be free to spread apart as long as they satisfy the binary CL objective. This can lead to \emph{over-dispersion}, where nearby pixels develop large feature differences, even if they belong to the same structure or object in the image. Mathematically, this means that for neighboring pixels $j\in\mathcal{N}(i)$, their feature similarity can vary widely: 
\begin{equation}
\langle f(x)_i,f(x)_j\rangle \rightarrow \text{anywhere within} [-\Delta,\Delta],
\end{equation}
since no term in binary CL enforces a small difference. This means binary CL lacks a mechanism to prevent large local variations, the feature differences $\delta_{BCL}$ can span the entire range of possible feature values $\Delta$. Thus, in the worst case:
\begin{equation}
\delta_{BCL}\approx\Delta.
\end{equation}
This means adjacent pixel features could be as different as features from completely unrelated image regions, which is undesirable for pixel-wise tasks like segmentation or medical image analysis. Therefore, the generalization error satisfies in the binary CL is
\begin{equation}\label{equ:bcl_bond}
R_{BCL}(f)\leq\hat{R}(f)+2\mathfrak{R}_n(\mathcal{F})+O\left(\sqrt{\frac{\log(1/\varepsilon)}{nN}}\right),
\end{equation}
causing the over-dispersion problem and limiting the generalization.

\subsection{Vector Contrastive Learning}
\paragraph{Derivation for Vector Contrastive Learning} Vector CL introduces an additional regression loss that explicitly enforces the mapping between feature distances and corresponding spatial displacement vectors. Its loss is
\begin{equation}
  \mathcal{L}_{\text{vec}} = \left\| v - \mathcal{V}(d') \right\|,
\end{equation}
where $v$ represents the ground-truth displacement vector and $\mathcal{V}(d')$ is a mapping function applied on the computed feature distances $d'$. It is formulated as a weighted sum of a vector template matrix $\mathbb{V}$, i.e., $\mathcal{V}(d')=\sum_{j=0}^{\mathcal{N}(i)}\mathbb{V}^{j}d'^{j}$, where the $d'^{j}=\frac{e^{\langle f(x)_i, f(x)_j \rangle/\tau}}{\sum_{j}^{\mathcal{N}(i)}e^{\langle f(x)_i, f(x)_j \rangle/\tau}}$. Therefore, the loss in vector CL $\mathcal{L}_{VCL}$ can be further formulated as
\begin{equation}
  \mathcal{L}_{VCL}=\begin{matrix}\left\|v-\sum_{j=0}^{\mathcal{N}(i)}\mathbb{V}^{j}\frac{e^{\langle f(x)_i, f(x)_j \rangle/\tau}}{\sum_{j}^{\mathcal{N}(i)}e^{\langle f(x)_i. f(x)_j \rangle/\tau}}\right\|\end{matrix}
\end{equation}
To minimize the $\mathcal{L}_{VCL}$, the model must adjust feature similarity $\langle f(x)_i, f(x)_j \rangle$ so that the weighted sum of $\mathbb{V}^{j}$ matches $v$. We assume that the $v$ can be modeled as 
\begin{equation}
v=\sum_{j}^{\mathcal{N}(i)}\alpha_{j}\mathbb{V}^{j}\ (\alpha_{j}\geq0, \sum_{j}^{\mathcal{N}(i)}\alpha_{j}=1),
\end{equation}
the model enforces $\frac{e^{\langle f(x)_{i},f(x)_{j}\rangle}/\tau}{Z}\approx\alpha_{j}$, where the $Z=\sum_{j}^{\mathcal{N}(i)}e^{\langle f(x)_i,f(x)_j\rangle}$. Therefore, the similarity $\langle f(x_i),f(x_j)\rangle$ can be further approximated as 
\begin{equation}
\langle f(x_i),f(x_j)\rangle\approx\tau\log\alpha_{j}+\tau\log Z.
\end{equation}
The maximum difference of our vector CL between any two pixels in the $\mathcal{N}(i)$ can be formulated is:
\begin{align}
\delta_{VCL}&=\max_{i,k}|\langle f(x_i),f(x_j)\rangle-\langle f(x_i),f(x_k)\rangle|\notag\\
&=\max_{i,k}|\tau\log\alpha_{j}+\tau\log Z-\tau\log\alpha_{k}-\tau\log Z|\notag\\
&=\tau\max_{i,k}|\log\frac{\alpha_j}{\alpha_k}|.
\end{align}
Owing to the $\alpha_{j},\alpha_{k}\in[\alpha_{\min},1]\ (\alpha_{\min}>0)$, when $\alpha_{j}=1, \alpha(k)=\alpha_{\min}$ the $\delta_{VCL}$ will be:
\begin{equation}\label{equ:detavcl}
\delta_{VCL}\leq\tau\log\frac{1}{\alpha_{\min}}.
\end{equation}
due to the normalization of weights ($\sum_{j}\alpha_{j}=1$), only when $\alpha_{j}=1$, all other $\alpha\in\mathcal{N}(i)$ are $0$. This situation where one weight is exactly 1 and all others are zero is degenerate (and would essentially revert to binary CL), so typically the effective $\delta_{VCL}$ is much smaller than the worst-case bound, i.e., $\delta_{VCL}\leq\tau\log\frac{1}{\alpha_{\min}}\ll\Delta$.

\paragraph{Generalization Bound for Vector Contrastive Learning} Recalling the generalization error bound for any $f\in\mathcal{F}$ in Equ.\ref{equ:gen}, we substitute the improved local dispersion bound $\delta_{VCL}$ for $\delta$ in the vector CL case and use the fact in Equ.\ref{equ:detavcl}, so that we will obtain:
\begin{equation}
  \begin{matrix}R_{VCL}(f)\le\hat{R}(f)+2\left(\frac{\tau\log\frac{1}{\alpha_{\min}}}{\Delta}\right)\mathfrak{R}_n(\mathcal{F})+O\!\left(\sqrt{\frac{\log(1/\varepsilon)}{nN}}\right)\end{matrix}
\end{equation}
Since typically $\tau\log(1/\alpha_{\min})\ll\Delta$ (because the normalization forces the weights to be spread across several pixels rather than concentrating on one), the factor $\frac{\tau\log\frac{1}{\alpha_{\min}}}{\Delta}$ is much smaller than 1. In contrast, in the binary CL setting, the worst-case dispersion is $\delta_{BCL}\approx\Delta$, leading to a much looser bound. Compared with the Equ.\ref{equ:bcl_bond}, our vector CL making a tighter generalization bound.

\section{Technical Details}

\subsection{\texorpdfstring{Appearance transformation operation $\mathcal{T}_{ap}$}{Details of appearance transformation}}
The appearance transformation operation $\mathcal{T}_{ap}$ consists of random noise, blurring, contrast, brightness, and in-painting. Specifically, 
\begin{compactenum}
\item For the random noise operation, it randomly generates Gaussian noise from the Gaussian distribution with $\mu=0$ (mean) and $\sigma\in[0,0.02]$ (variance), and adds it to the image.
\item For the random blurring, it performs random Gaussian blurring with $\mu=0$ and $\sigma\in[0,0.05]$ on the image.
\item For the random contrast, it performs multiplicative transformation with $(x-\bar{x})*\gamma+\bar{x}$ on the images $x$, where the $\bar{x}$ is the average value and the $\gamma\in[0.5,1.5]$ is the scaling value.
\item For the random brightness, it performs additive transformation with $x+\beta$ on the images, where the $\beta$ is randomly sampled from a Gaussian distribution with $\mu=0$ and $\sigma\in[0,0.1]$.
\item For the random in-painting, it randomly selects boxes, and the contents of these regions are replaced by the noise from a uniform distribution.
\end{compactenum}
These five sub-operations are used sequentially with a probability of 0.9 for the image with appearance transformation.

\subsection{\texorpdfstring{Space transformation operation $\mathcal{T}_{sp}$}{Details of space transformation}}
We utilize random affine transformation \cite{brown1992survey} to construct the space transformation operation $\mathcal{T}_{sp}$. For a clearer illustration, we introduce it in 2D situation. As shown in Equ.\ref{equ:def}, it as four sub-operations including the translation $t_{x},t_{y}\in[-0.2,0.2]$, rotation $\theta\in[-\pi/9,\pi/9]$, shearing $sh_{x},sh_{y}\in[-\pi/32,\pi/32]$, and scaling $s_x,s_y\in[0.5,1.5]$, where the $H,W$ in the Equ.\ref{equ:def} are the height and width of the medical images. These operations are These operations form an affine transformation matrix $\phi_{ab}$ that transforms the position of each pixel $\textbf{\text{p}}_{a}$ in the image to a new position $\textbf{\text{p}}_{ab}$. Therefore, the ground truth vector $v^{i}_{ab}$ will be generated via the coordinate difference on the image grid $v^{i}_{ab}=\textbf{\text{p}}^{i}_{ab}-\textbf{\text{p}}^{i}_{a}$, and the vectors at each position jointly construct the displacement vector field (DVF) $\psi_{ab}=\{v^{i}_{ab}\}_{i\in\mathbb{R}}$. The space-transformed view $x_{b}$ is generated by the DVF $\psi_{ab}$ via moving the pixels to target positions and completing the non-integer coordinates by bilinear interpolation.
\begin{table*}
  \centering
\begin{align}
&
\begin{array}{c}\label{equ:def}
  \phi_{ab}=
\overbrace{\begin{bmatrix}
1 & 0 & t_{x}H \\
0 & 1 & t_{y}W \\
0 & 0 & 1
\end{bmatrix}}^{\textbf{\text{Translation}}}
\overbrace{\begin{bmatrix}
\cos\theta & -\sin\theta & 0 \\
\sin\theta & \cos\theta & 0 \\
0 & 0 & 1
\end{bmatrix}}^{\textbf{\text{Rotation}}}
\overbrace{\begin{bmatrix}
1 & sh_{x} & 0 \\
sh_{y} & 1 & 0 \\
0 & 0 & 1
\end{bmatrix}}^{\textbf{\text{Shearing}}}
\overbrace{\begin{bmatrix}
s_{x} & 0 & 0 \\
0 & s_{y} & 0 \\
0 & 0 & 1
\end{bmatrix}}^{\textbf{\text{Scaling}}}=
\begin{bmatrix}
a_{11} & a_{12} & t_{x}H \\
a_{21} & a_{22} & t_{y}W \\
0 & 0 & 1
\end{bmatrix}\\=
\begin{bmatrix}
s_{x}\cos\theta+sh_{x}s_{y}\sin\theta & -s_{x}\sin\theta+sh_{x}s_{y}\cos\theta & t_{x}H \\
sh_{y}s_{x}\cos\theta & -sh_{y}s_{x}\sin\theta+s_{y}\cos\theta & t_{y}W \\
0 & 0 & 1
\end{bmatrix}
,
\end{array}\\
&
\begin{array}{c}
v^{i}_{ab}=\textbf{\text{p}}^{i}_{ab}-\textbf{\text{p}}^{i}_{a}=\phi_{ab}\cdot
\begin{bmatrix}
p_{x}^{i} \\
p_{y}^{i} \\
1 \\
\end{bmatrix}-
\begin{bmatrix}
p_{x}^{i} \\
p_{y}^{i} \\
1 \\
\end{bmatrix}=
\begin{bmatrix}
(a_{11}-1)p_{x}^{i}+a_{12}p_{y}^{i}+t_{x}H \\
a_{21}p_{x}^{i}+(a_{22}-1)p_{y}^{i}+t_{y}W \\
1 \\
\end{bmatrix}=
\begin{bmatrix}
v^{i}_{x} \\
v^{i}_{y} \\
1 \\
\end{bmatrix}, 
\psi_{ab}=\{v^{i}_{ab}\}_{i\in\mathbb{R}}\notag
\end{array}
\end{align}
\end{table*}
\subsection{\texorpdfstring{Mask $\epsilon_{ab}$ indicates matched regions}{Mask}}
The mask $\epsilon_{ab}$ eliminates the content mismatch of two views $x_{a}$, $x_{b}$ caused by the spatial transformation. It is calculated from the DVF $\psi_{ab}$ according to whether the transformed coordinates exceed the image grid. For each position $i$ on image grid, the value of the $\epsilon^{i}_{ab}$ is
\begin{equation}\label{equ:eps}
\epsilon_{ab}^{i}=\textbf{\text{1}}_{(0,H)}(v^{i}_{x}+p^{i}_{x})\wedge\textbf{\text{1}}_{0,W}(v^{i}_{y}+p^{i}_{y}),
\end{equation}
where the $\textbf{\text{1}}_{(0,H)}(\cdot)$ and $\textbf{\text{1}}_{(0,W)}(\cdot)$ are Iverson Notation \cite{iverson1962programming}. Therefore, the mask $\epsilon_{ab}$ be generated whose 0 values indicate the mismatched regions and 1 values indicate the matched regions.

\subsection{\texorpdfstring{Multi-scale fusion operation $\bigodot$ in our VPA}{Details of multi-scale fusion operation}}
The fusion operation $\bigodot$ in our vector pyramid aggregation integrates the DVFs from different levels. Here, we utilize the fusion of DVFs in level 0 $\psi'^{0}_{ab}$ and level 1 $\psi'^{1}_{ab}$ to introduce this operation. It has three steps: \textbf{1) Scale alignment:} The level 0 DVF $\psi'^{0}_{ab}$ is up-sampled to the same size of level 1 DVF $\psi'^{1}_{ab}$ via bilinear interpolation. Then the values are enlarged (double) to adapt to the size of the level 1 grid. \textbf{2) Space alignment:} To align the center coordinates of the vectors in these two level DVFs, the level 0 DVF is transformed to align the level 1 DVF. \textbf{3) Vector fusion:} Finally, the DVFs in two levels are fused via addition. The whole fusion operation $\bigodot$ is formulated as, 
\begin{equation}\label{equ:fusion}
  \psi'^{1}_{ab}\bigodot\psi'^{0}_{ab}=\overbrace{\psi'^{1}_{ab}+\underbrace{\psi'^{1}_{ab}(\overbrace{2*\mathcal{I}_{H\times W}(\psi'^{0}_{ab})}^{\text{Scale alignment}})}_{\text{Space alignment}}}^{\text{Vector fusion}},
\end{equation}
where $H$ and $W$ are the height and width of level 1 grid, and $\mathcal{I}$ is the bilinear interpolation.

\subsection{\texorpdfstring{Vector template $\mathbb{V}^{N\times N}$ maps distances as vector}{Vector template matrix}}
\begin{figure}
    \centering
    \includegraphics[width=\linewidth]{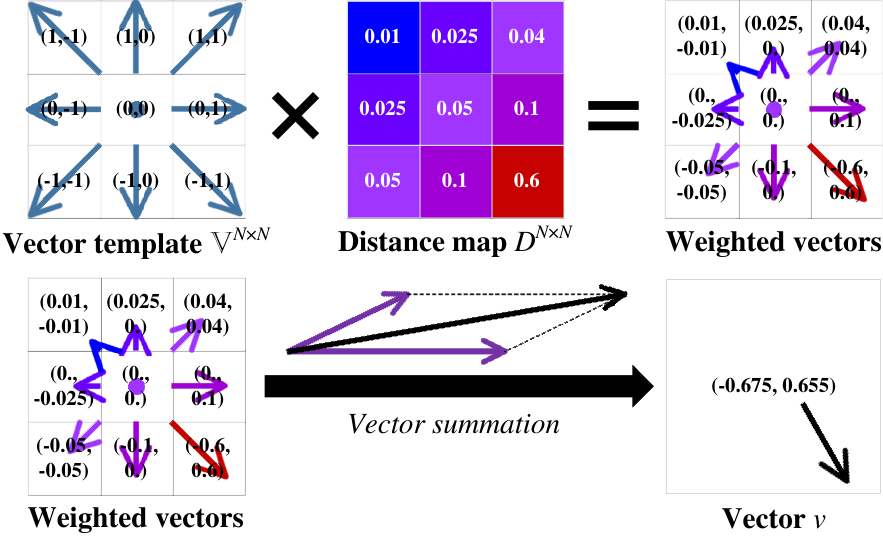}
    \caption{The details of our vector template matrix. It is multiplied to the distance map mapping the distances as vectors. The values in the distance map are examples.}
    \label{fig:vt}
\end{figure}

The vector template matrix describes the basic spatial relationship of pixels that indicates the vectors in which the center coordinates pointing to the coordinates in the receptive field. As shown in Fig.\ref{fig:vt}, for a clear illustration, we assume that the template is a 2D matrix (2 channels) with 3$\times$3 receptive fields. The value in each position is the vector, i.e., $(x,y)$, that indicates the relative position offset from the center coordinate so that the $\mathbb{V}^{3\times 3}=\{(x,y)|x,y\in\{-1,0,1\}\}$. To map the distances as vectors, the vector template matrix is multiplied by the distance map $D^{N\times N}$ of the receptive field corresponding to the center coordinate, i.e., $D^{3\times 3}\cdot\mathbb{V}^{3\times 3}=\{(x,y)|x=\mathbb{V}^{i,j}_{[0]}*D^{i,j},y=\mathbb{V}^{i,j}_{[1]}*D^{i,j},i,j\in\{0,1,2\}\}$, for weighted vectors in a matrix. Finally, these weighted vectors are summed for the vector $v=\sum_{i,j}^{j,j\in\{0,1,2\}}D^{i,j}\cdot \mathbb{V}^{i,j}$ indicating the correspondence between the center coordinate $(0,0)$ and the coordinates in the field.

\section{Experiment Details}
\subsection{Datasets}
\begin{table}
  \centering
\resizebox{\linewidth}{!}{
  \begin{tabular}{lccccccccc}
  \textbf{Dataset}                       &\textbf{Type}    &\textbf{Num}  &\textbf{D}       &\textbf{P} &\textbf{Task}\\
  \hline
  CANDI \cite{kennedy2012candishare}     &3D T1 brain MRI     &103           &$\surd$          &           &S\\
  FeTA21 \cite{payette2023fetal}         &3D T2 brain MRI     &80            &$\surd$          &           &S\\
  SCR \cite{van2006segmentation}         &2D chest X-ray   &247           &$\surd$          &           &S\\
  KiPA22 \cite{he2021meta}               &3D kidney CT     &130           &$\surd$          &           &S\\
  FIVES \cite{jin2022fives}              &2D fundus        &800           &$\surd$          &           &S\\
  PDCXR \cite{kermany2018identifying}    &2D chest X-ray   &5,956         &$\surd$          &           &C\\
  STOIC \cite{revel2021study}            &3D chest CT      &2,000         &$\surd$          &           &C\\
  ChestX-ray8 \cite{wang2017chestx}      &2D chest X-ray   &112,120       &                 &$\surd$    &-\\
  PPMI (T1) \cite{marek2011parkinson}    &3D T1 brain MRI  &837           &                 &$\surd$    &-\\
  \end{tabular}}
  \caption{A Total of 9 publicly available datasets are involved in this paper for the experiments, achieving great reproducibility. The ``D" and ``P" mean the datasets are used for downstream tasks and pretraining tasks. The ``S" and ``C" are the segmentation and classification tasks.}\label{tab:data}
\end{table}
As shown in Tab.\ref{tab:data}, eight publicly available datasets are involved in this paper, specifically,

\textbf{CANDI \cite{kennedy2012candishare}} The Child and Adolescent NeuroDevelopment Initiative (CANDI) dataset has 103 T1 brain MR volumes from 57 males and 46 females. Totally 28 brain tissue regions are annotated for masks. For the segmentation task (CANDI$_{S}$), 40, 20, and 43 volumes are used as training, validation, and test sets. Following \cite{he2022learning}, we resize and crop 160$\times$160$\times$128 volumes on the brain regions, and then normalize the intensity via $\frac{x-\min(x)}{\max(x)-\min(x)}$.

\textbf{FeTA21 \cite{payette2023fetal}} The Fetal Tissue Annotation 2021 (FeTA21) challenge dataset has 120 fetal T2w brain MR volumes, and 80 of them are available as the training data in the challenge. We split the 80 volumes and set 20 of them as the training set, 20 of them as the validation set, and 40 of them as test set. We normalize the intensity of the images via $\frac{x-\min(x)}{\max(x)-\min(x)}$ and use the $128\times128\times128$ random cropping to unify the input size.

\textbf{SCR \cite{van2006segmentation}} The segmentation of chest radiographs (SCR) dataset is from the JSRT database \cite{shiraishi2000development} with 247 2048$\times$2048 posterior to anterior (PA) chest radiographs. Three chest-related structures, including the heart, chest, and clavicle, are annotated for masks. We set 100 of them as the training set, 47 of them as a validation set, and 100 of them as the test set. Following \cite{he2022learning}, we resize the images to 512$\times$512, and normalize the intensity via $x/255$ for the segmentation task (SCR$_{S}$).

\textbf{KiPA22 \cite{he2021meta}} The kidney parsing 2022 (KiPA22) challenge dataset has 130 kidney CT volumes. These images are cropped from 130 abdominal CT angiography volumes for kidney regions with tumors. Four kidney-related structures, including the kidney, vessel, vein, and tumor, are annotated for masks. In 2D evaluation (KiPA$^{2D}_{S}$), 13,846 2D slices from 70 volumes are used as the training set, 5,864 2D slices from 30 volumes are used as the validation set, 5,959 2D slices from 30 volumes are used as the test set. We normalize the intensity of the images via $\frac{\max(\min(0,x),2048)}{2048}$, and use the $128\times128$ randomly to unify the input size. For 3D evaluation (KiPA22$^{3D}_{S}$), 70, 30, and 30 volumes are used as training, validation, and test sets. We normalize the intensity of the images via $\frac{\max(\min(0,x),2048)}{2048}$, and use the $128\times128\times128$ randomly to unify the input size.

\textbf{FIVES \cite{jin2022fives}} The fundus image vessel segmentation (FIVES) dataset consists of 800 color fundus photographs with vessel annotation from 573 patients. 540, 60, and 200 of the images are used as training, validation, and test sets. We resize the images to $512\times512$, and normalize the intensity via $x/255$ for the segmentation task (FIVES$_{S}$).

\textbf{PDCXR \cite{kermany2018identifying}} The pneumonia Detection Using Chest X-ray (PDCXR) dataset has 5,856 chest X-ray images for the diagnosis of pneumonia. Following \cite{kermany2018identifying}, 3,659 of them are used as training set (2,714 pneumonia, 945 normal), 1,573 of them are used as validation set (1,169 pneumonia, 404 normal), and 624 of them are used as test set (390 pneumonia, 234 normal). We resize the images to $512\times512$, and normalize the intensity via $x/255$ for classification task (PDCXR$_C$).

\textbf{STOIC \cite{revel2021study}} The STOIC dataset is from the STOIC 2021 challenge with 2000 chest CT volumes for COVID-19 diagnosis. Following \cite{He_2023_CVPR}, 1000 of them are used as training set (603 COVID-19, 397 normal), 400 of them are used as validation set (241 COVID-19, 159 normal), and 600 of them are used as test set (361 COVID-19, 239 normal). We utilize the Lungmask \cite{hofmanninger2020automatic} to extract the lung regions avoiding the interference of the background, resample the resolutions to 1$mm^{3}$, and normalize the intensity via $\frac{\max(\min(0,x),2048)}{2048}$ for classification task (STOIC$_{C}$).

\textbf{ChestX-ray8 \cite{wang2017chestx}} The ChestX-ray8 is our pre-training dataset in 2D evaluation. It has 112,120 frontal-view chest X-ray images with 1024$\times$1024 resolution. 44,810 of them are scanned from the anterior to posterior (AP) view and 67,310 of them are scanned from the PA view. We resize the images to 512$\times$512, and normalize the intensity $x/255$. During the pre-training, 384$\times$384 patches are randomly cropped for augmentation.

\textbf{PPMI \cite{marek2011parkinson}} The PPMI is our pre-training dataset in 3D evaluation. It is extracted from the PPMI database which is a large Parkinson progression marker initiative database, for 837 T1 brain MR volumes. Following the pre-processing in the CANIA dataset, we resize and crop 160$\times$160$\times$128 volumes on the brain regions, and then normalize the intensity via $\frac{x-\min(x)}{\max(x)-\min(x)}$. We also extract the brain regions via HD-BET \cite{isensee2019automated} to avoid the interference of background. During the pre-training, 128$\times$128$\times$128 patches are randomly cropped for augmentation.

\subsection{Implementations}
\begin{figure}
    \centering
    \includegraphics[width=\linewidth]{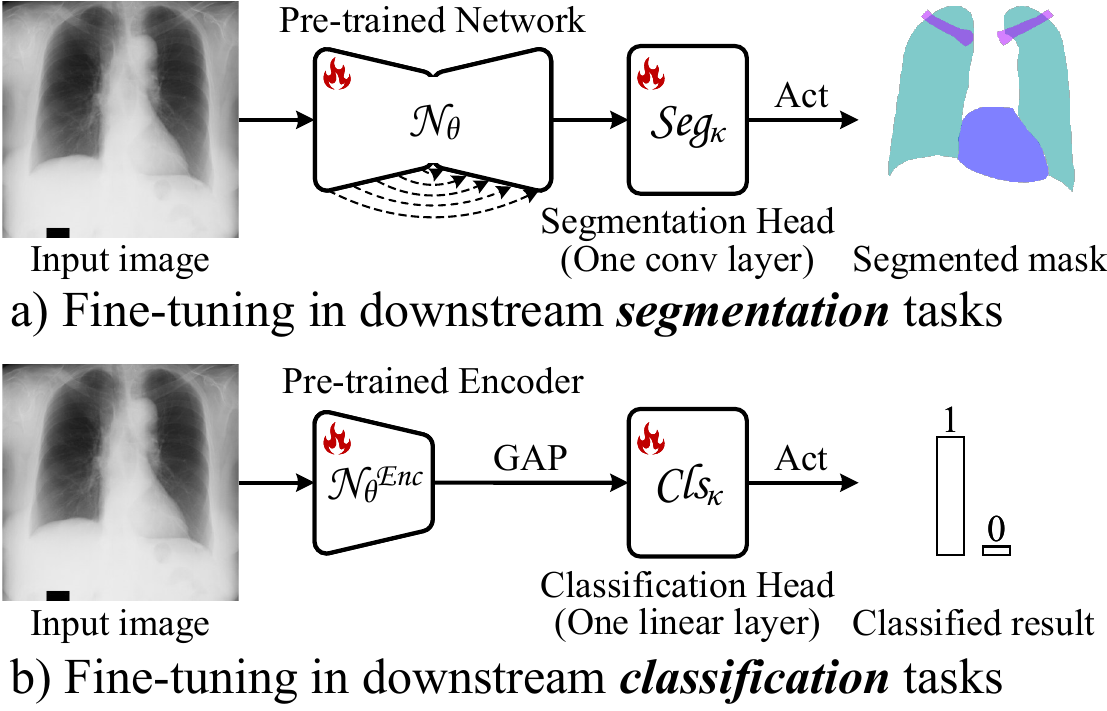}
    \caption{The detailed implementations in our downstream tasks, including the a) segmentation and b) classification.}
    \label{fig:downstream}
\end{figure}
Our experiments are implemented by PyTorch \cite{paszke2019pytorch} which is a widely recognized deep learning library on NVIDIA A100 SXUM4 GPU with 40 GB memory. Specifically,
\subsubsection{Pre-training implementation}
In both 2D evaluation and 3D evaluation, we utilize the Adam optimizer \cite{kingma2014adam} with the learning rate of $1\times10^{-4}$ and iterations of $2\times10^{5}$. For 2D evaluation, 24 2D chest X-ray images from ChestX-ray8 \cite{wang2017chestx} dataset are randomly sampled as a batch in each iteration with 384$\times$384 random cropping augmentation. For 3D evaluation, 2 3D brain MR volumes from the PPMI \cite{marek2011parkinson} dataset are randomly sampled as a batch in each iteration with 128$\times$128$\times$128 random cropping augmentation.

\subsubsection{Downstream implementation}
As shown in Fig.\ref{fig:downstream}, following \cite{He_2023_CVPR}, we utilize the fine-tuning evaluation to demonstrate the adaptation ability in downstream tasks. The gradient optimizes all parameters through the frameworks during the training. All the downstream tasks are trained by Adam optimizer \cite{kingma2014adam} with the learning rate of $1\times10^{-4}$ and iterations of $4\times10^{4}$.

\textbf{For segmentation tasks}, the whole pre-trained network $\mathcal{N}_{\theta}$ is utilized to extract the pixel-wise features (level 4) $f^{4}$, and these features are putted into a new segmentation head $Seg_{\kappa}$ to predict the final segmentation masks. The segmentation head is a convolutional layer to map the features to the channels of segmentation targets, thus constructing a segmentation framework for the SCR$_{S}$, KiPA22$^{2D}_{S}$, FIVES$_{S}$, CANDI$_{S}$, and KiPA22$^{3D}_{S}$ tasks. We use the sum of Dice loss \cite{zhao2020rethinking} and cross-entropy loss \cite{mao2023cross} between the predicted masks and ground truths to train the segmentation tasks.

\textbf{For classification tasks}, the encoder part of the pre-trained network $\mathcal{N}^{Enc}_{\theta}$ is utilized to extract the high-level features (level 0) $f^{0}$, and a global adaptive pooling is used to compress features. Then these compressed features are putted into a classification head $Cls_{\kappa}$ to achieve the classified results, thus constructing a segmentation framework for the PDCXR$_{C}$ and STOIC$_{C}$ tasks. We utilize the cross-entropy loss \cite{mao2023cross} between the predicted categories and ground truths to train the classification.

\section{More Framework Analysis and Results}
\begin{table}
\centering
\resizebox{\linewidth}{!}
{
\begin{tabular}{ll|cc|cc|c}
Backbone
&Methods
&\textbf{SCR}$^{25\%}_{S}$
&\textbf{PDCXR}$_{C}$
&\textbf{KiPA22}$^{2D}_{S}$
&\textbf{FIVES}$_{S}$
&\textbf{AVG}
\\
\hline
U-Net \cite{ronneberger2015u}
&\color{gray}Scratch
&\color{gray}81.8
&\color{gray}90.4
&\color{gray}74.1
&\color{gray}79.4
&\color{gray}81.4
\\
& \textbf{COVER}
&94.0$\color{blue}_{(+12.2)}$
&95.9$\color{blue}_{(+5.5)}$
&80.0$\color{blue}_{(+5.9)}$
&87.2$\color{blue}_{(+7.8)}$
&89.3$\color{blue}_{(+7.9)}$
\\
\cdashline{1-6}[0.8pt/2pt]
TransUNet \cite{chen2024transunet}
& \color{gray}Scratch
& \color{gray}89.2
& \color{gray}77.4
& \color{gray}57.6
& \color{gray}81.7
& \color{gray}76.5
\\
& \textbf{COVER}
& 93.4$\color{blue}_{(+4.2)}$
& 90.6$\color{blue}_{(+13.2)}$
& 76.7$\color{blue}_{(+19.1)}$
& 86.5$\color{blue}_{(+4.8)}$
& 86.8$\color{blue}_{(+10.3)}$
\\
\cdashline{1-6}[0.8pt/2pt]
SwinUNet \cite{cao2022swin}
& Scratch
& \color{gray}85.6
& \color{gray}93.0
& \color{gray}69.1
& \color{gray}77.0
& \color{gray}81.2
\\
& \textbf{COVER}
& 86.9$\color{blue}_{(+1.3)}$
& 95.6$\color{blue}_{(+2.6)}$
& 72.6$\color{blue}_{(+3.5)}$
& 84.8$\color{blue}_{(+7.8)}$
& 85.0$\color{blue}_{(+3.8)}$
\\
\cdashline{1-6}[0.8pt/2pt]
U-KAN \cite{li2024ukan}
& \color{gray}Scratch
& \color{gray}89.4
& \color{gray}89.3
& \color{gray}61.3
& \color{gray}80.3
& \color{gray}80.1
\\
& \textbf{COVER}
& 93.8$\color{blue}_{(+4.4)}$
& 95.6$\color{blue}_{(+6.3)}$
& 71.8$\color{blue}_{(+10.5)}$
& 84.9$\color{blue}_{(+4.6)}$
& 86.5$\color{blue}_{(+6.4)}$
\end{tabular}
}
\caption{Our COVER has great cross-architecture compatibility that achieves significant improvement on all U-Net, TransUNet, SwinUNet, and UKAN. We evaluate it in our 2D setting.}
\label{tab:archi}
\end{table}
\subsection{Analysis of the reliability}

\begin{table}[tb]
\centering
\resizebox{\linewidth}{!}
{
\begin{tabular}{ccccccccccccccc}
  \multicolumn{2}{c}{\textbf{SCR$^{25\%}_{S}$}}
  &\multicolumn{2}{c}{\textbf{PDCXR$_{C}$}}
  &\multicolumn{2}{c}{\textbf{KiPA22$^{2D}_{S}$}}
  &\multicolumn{2}{c}{\textbf{FIVES$_{S}$}}
  &\multicolumn{2}{c}{\textbf{AVG}}
  \\
  \emph{Cor} $\uparrow$
  &\emph{p} $\downarrow$
  &\emph{Cor} $\uparrow$
  &\emph{p} $\downarrow$
  &\emph{Cor} $\uparrow$
  &\emph{p} $\downarrow$
  &\emph{Cor} $\uparrow$
  &\emph{p} $\downarrow$
  &\emph{Cor} $\uparrow$
  &\emph{p} $\downarrow$
  \\
  \hline
  0.908
  & $<$0.001
  & 0.898
  & $<0.001$
  & 0.949
  & $<$0.001
  & 0.995
  & $<$0.001
  & 0.938
  & $<0.001$
\end{tabular}
}
\caption{The test-retest reliability analysis \cite{guttman1945basis} of our COVER on the tasks of 2D evaluation. The \emph{Cor} is the Pearson correlation coefficient \cite{cohen2009pearson}, and the \emph{p} is the p-value.}
\label{tab:reliability}
\end{table}
As shown in Tab.\ref{tab:reliability}, we utilize the test-retest analysis \cite{guttman1945basis} to evaluate the reliability of our COVER. We pre-trained and adapted our method twice in our 2D evaluation from different initialization states, and then calculated the correlation coefficient \cite{cohen2009pearson} and p-value between these two results. Our COVER achieved 0.938 average Cors over four tasks demonstrating very high consistency between two training sessions. All p-values lower than 0.001 illustrated the significant consistency. Therefore, these results show our powerful reliability across initialization states, supporting the implementation in the application.

\subsection{Analysis of cross-architecture compatibility}

As shown in Tab.\ref{tab:archi}, our COVER has great cross-architecture compatibility that achieves significant improvement on all U-Net \cite{ronneberger2015u} (CNN-based), TransUNet \cite{chen2024transunet} (CNN-Transformer-based), SwinUNet \cite{cao2022swin} (Transformer-based), and UKAN \cite{li2024ukan} (KAN-based). We utilize these four networks with different paradigms as the backbone network in our COVER framework and train the framework on our 2D evaluation setting. Compared with the ``scratch"  on these networks, our COVER has achieved more than 3\% average improvement owing to the learned knowledge from the pre-training data. Especially, on the TransUNet which utilizes a vision transformer and is easy to fall into an over-fitting state, our COVER brings a significant 10.3\% average improvement.

\subsection{Analysis pre-training data amount}
\begin{figure}
    \centering
    \includegraphics[width=\linewidth]{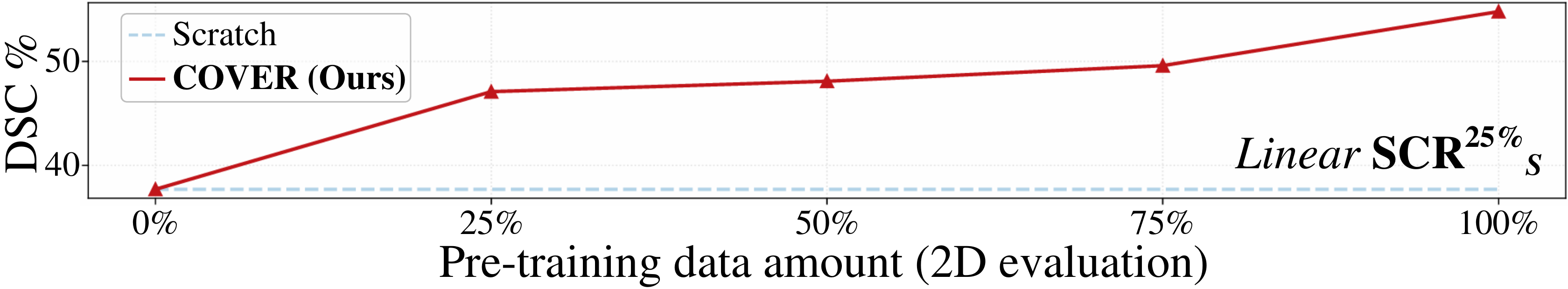}
    \caption{The analysis of the pre-training data amount. When enlarging the pre-training data amount, the performance of our COVER is improved gradually.}
    \label{fig:pre_data}
\end{figure}
As shown in Fig.\ref{fig:pre_data}, we evaluate the variation of our COVER's performance with the enlarging of the pre-training data amount on our 2D evaluation and adapt the pre-trained model to the SCR$^{25\%}_{S}$ via linear evaluation. Compared with the ``scratch", our COVER will bring a significant improvement even though only 25\% pre-training data is involved. When further enlarging the -pre-training dataset, the gain of performance gradually decreases owing to the similarity of the same category of medical images in the pre-training dataset. Fortunately, our COVER has a powerful modeling ability for pixel-wise features, enabling the learning from the details effectively, so the performance of the model is still improving gradually.

\subsection{Visualization of the segmentation results}

\begin{figure*}
    \centering
    \includegraphics[width=\linewidth]{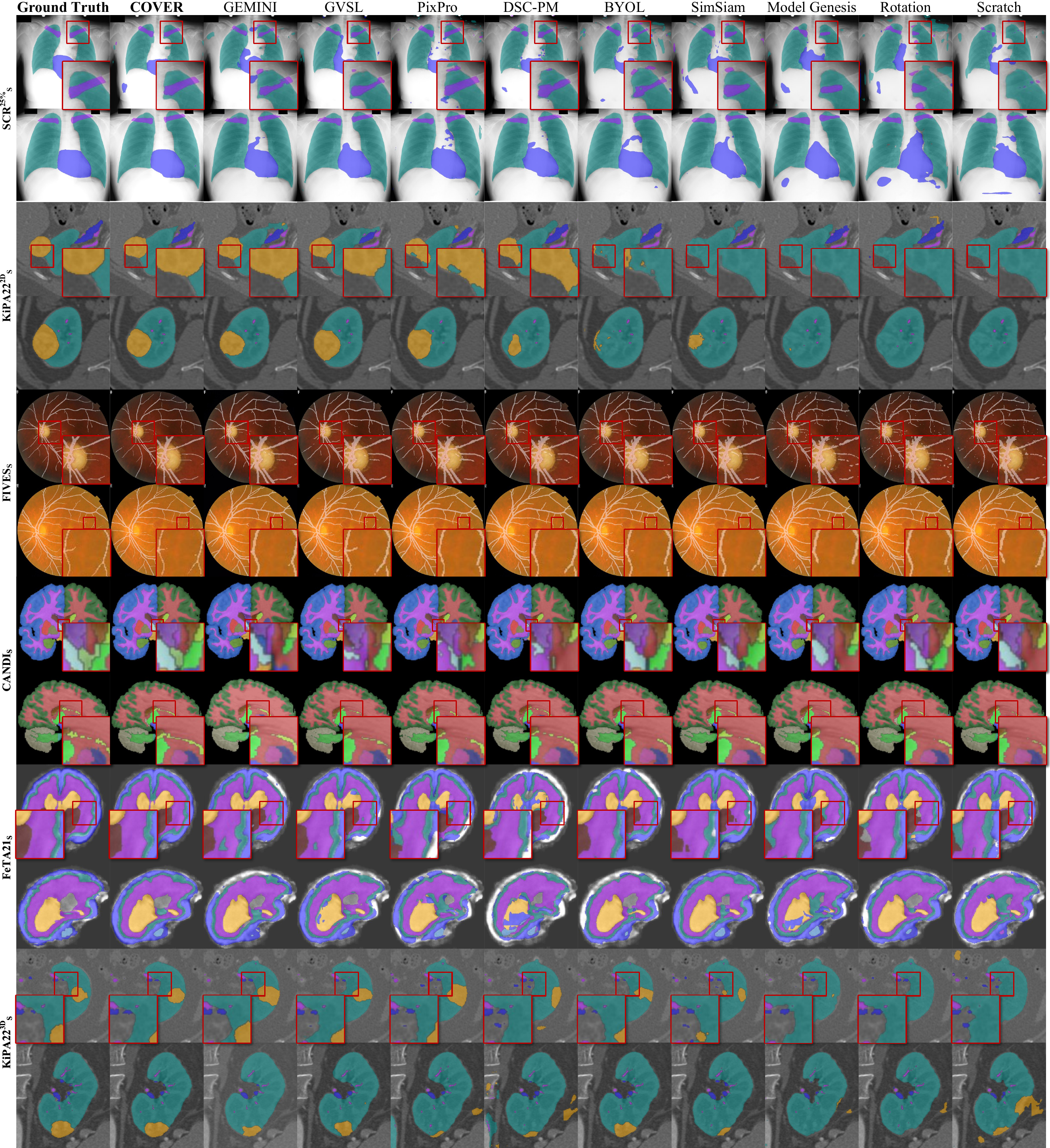}
    \caption{The visualization of the segmentation results for the methods with the top 2 average scores in each type. Our COVER achieves excellent integrity and fine segmentation ability for thin structures owing to its effective representation of pixel-wise features.}
    \label{fig:seg}
\end{figure*}
The visualization of the segmentation results (Fig.\ref{fig:seg}) demonstrates our superiority in the adaptation of pixel-wise tasks. Due to our vector CL with distance modeling, the pre-training enables the networks pixel-wise representation with controllable dispersion. It has two observations: 1) For large objects with varied appearances like the tumors in KiPA22$^{2D}_{S}$ and KiPA22$^{3D}_{S}$, our COVER achieves excellent integrity owing to its disentanglement of underlying explanatory factors hidden in low-level sensory data. The BYOL, SimSiam, Model Genesis, Rotation, and ``Scratch" have poor performance on the tumors, because of their lack of modeling for distinct features. 2) For small objects like the thin vessels in FIVE$_{S}$, small brain tissues in CANDI$_{S}$ and FeTA21 $_{S}$, and clavicles in SCR$^{25\%}_{S}$, our COVER also has fine segmentation owing to our pixel-wise representation with distance modeling. Such representation preserves the distinction of detail features thus making the networks easy to segment these regions in downstream tasks.

\subsection{Visualization of multi-scale vectors}
\begin{figure*}
    \centering
    \includegraphics[width=\linewidth]{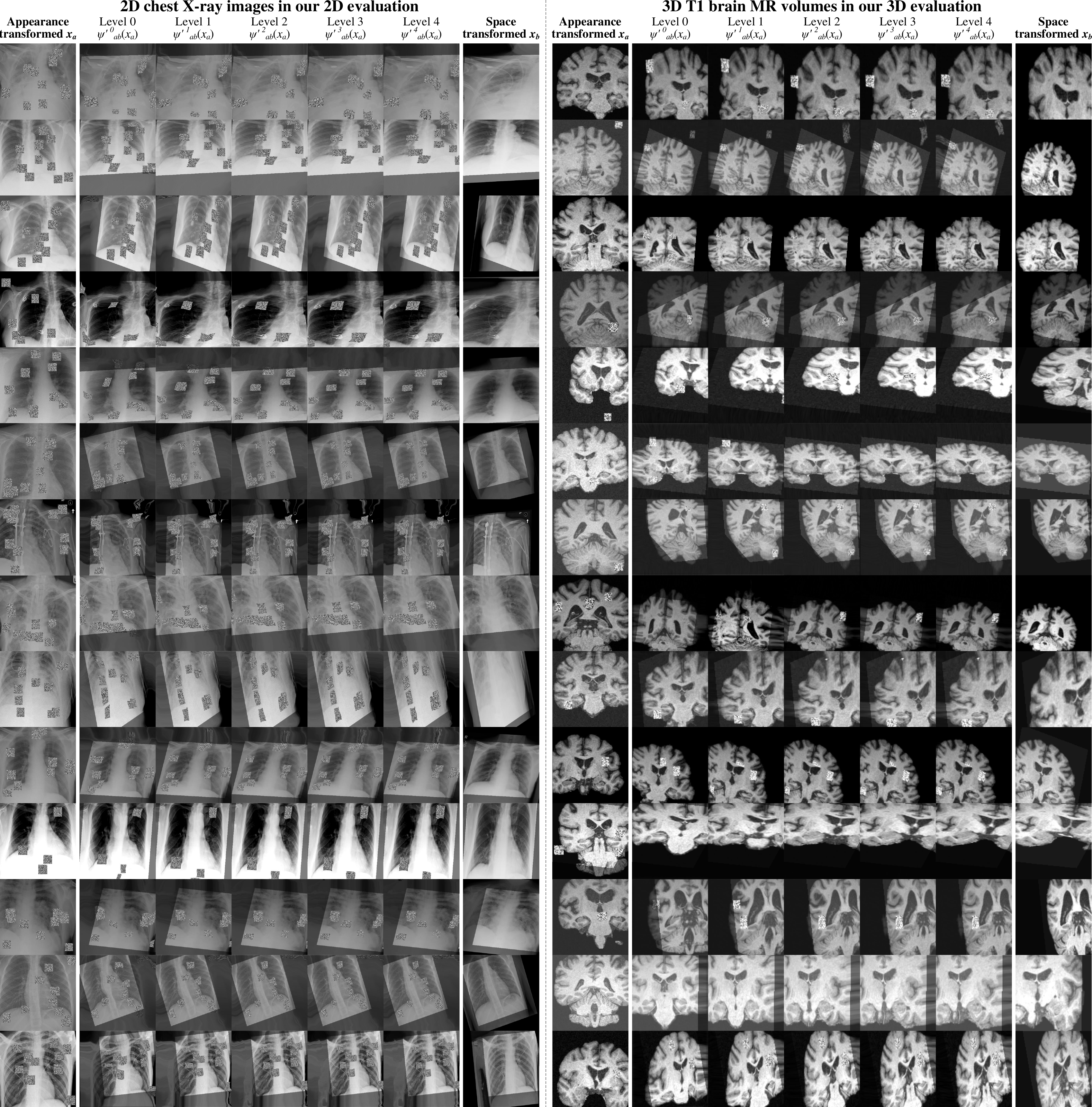}
    \caption{\textbf{Extension of multi-scale vectors analysis} (Fig.8 in manuscript): The visualization of the regressed vectors via the alignment of two views. Our VPA is able to discover the correspondence of the semantics at multiple levels for both global and detailed representation.}
    \label{fig:align}
\end{figure*}
We visualize more results of our regressed multi-scale vectors (corresponding to Fig.8 in the manuscript) in Fig.\ref{fig:align}. It aligns the appearance-transformed view $x_{a}$ to the space-transformed view $x_{b}$ via the deformation of regressed vectors in different scale levels. In a large number of cases, we can find that our VPA effectively aligns the regions with the same semantics between the views, illustrating that the regressed vectors are able to indicate the correspondence of the same semantics. Therefore, this means that in the embedding space, our vector CL can model closer distances between the same semantic objects and farther distances between different semantic objects, thus enabling our COVER to discover accurate correspondence in the receptive field.
{
    \small
    \bibliographystyle{ieeenat_fullname}
    \bibliography{main}

\begin{thebibliography}{87}
\providecommand{\natexlab}[1]{#1}
\providecommand{\url}[1]{\texttt{#1}}
\expandafter\ifx\csname urlstyle\endcsname\relax
  \providecommand{\doi}[1]{doi: #1}\else
  \providecommand{\doi}{doi: \begingroup \urlstyle{rm}\Url}\fi

\bibitem[Balakrishnan et~al.(2019)Balakrishnan, Zhao, Sabuncu, Guttag, and
  Dalca]{balakrishnan2019voxelmorph}
Guha Balakrishnan, Amy Zhao, Mert~R Sabuncu, John Guttag, and Adrian~V Dalca.
\newblock Voxelmorph: a learning framework for deformable medical image
  registration.
\newblock \emph{IEEE transactions on medical imaging}, 38\penalty0
  (8):\penalty0 1788--1800, 2019.

\bibitem[Bartlett and Mendelson(2002)]{bartlett2002rademacher}
Peter~L Bartlett and Shahar Mendelson.
\newblock Rademacher and gaussian complexities: Risk bounds and structural
  results.
\newblock \emph{Journal of Machine Learning Research}, 3\penalty0
  (Nov):\penalty0 463--482, 2002.

\bibitem[Bengio et~al.(2013)Bengio, Courville, and
  Vincent]{bengio2013representation}
Yoshua Bengio, Aaron Courville, and Pascal Vincent.
\newblock Representation learning: A review and new perspectives.
\newblock \emph{IEEE transactions on pattern analysis and machine
  intelligence}, 35\penalty0 (8):\penalty0 1798--1828, 2013.

\bibitem[Bommasani et~al.(2021)Bommasani, Hudson, Adeli, Altman, Arora, von
  Arx, Bernstein, Bohg, Bosselut, Brunskill,
  et~al.]{bommasani2021opportunities}
Rishi Bommasani, Drew~A Hudson, Ehsan Adeli, Russ Altman, Simran Arora, Sydney
  von Arx, Michael~S Bernstein, Jeannette Bohg, Antoine Bosselut, Emma
  Brunskill, et~al.
\newblock On the opportunities and risks of foundation models.
\newblock \emph{arXiv preprint arXiv:2108.07258}, 2021.

\bibitem[Brown(1992)]{brown1992survey}
Lisa~Gottesfeld Brown.
\newblock A survey of image registration techniques.
\newblock \emph{ACM computing surveys (CSUR)}, 24\penalty0 (4):\penalty0
  325--376, 1992.

\bibitem[Cao et~al.(2022)Cao, Wang, Chen, Jiang, Zhang, Tian, and
  Wang]{cao2022swin}
Hu Cao, Yueyue Wang, Joy Chen, Dongsheng Jiang, Xiaopeng Zhang, Qi Tian, and
  Manning Wang.
\newblock Swin-unet: Unet-like pure transformer for medical image segmentation.
\newblock In \emph{European conference on computer vision}, pages 205--218.
  Springer, 2022.

\bibitem[Caron et~al.(2018)Caron, Bojanowski, Joulin, and Douze]{caron2018deep}
Mathilde Caron, Piotr Bojanowski, Armand Joulin, and Matthijs Douze.
\newblock Deep clustering for unsupervised learning of visual features.
\newblock In \emph{European Conference on Computer Vision}, 2018.

\bibitem[Chaitanya et~al.(2020)Chaitanya, Erdil, Karani, and
  Konukoglu]{chaitanya2020contrastive}
Krishna Chaitanya, Ertunc Erdil, Neerav Karani, and Ender Konukoglu.
\newblock Contrastive learning of global and local features for medical image
  segmentation with limited annotations.
\newblock \emph{Advances in neural information processing systems},
  33:\penalty0 12546--12558, 2020.

\bibitem[Chen et~al.(2024{\natexlab{a}})Chen, Mei, Li, Lu, Yu, Wei, Luo, Xie,
  Adeli, Wang, et~al.]{chen2024transunet}
Jieneng Chen, Jieru Mei, Xianhang Li, Yongyi Lu, Qihang Yu, Qingyue Wei,
  Xiangde Luo, Yutong Xie, Ehsan Adeli, Yan Wang, et~al.
\newblock Transunet: Rethinking the u-net architecture design for medical image
  segmentation through the lens of transformers.
\newblock \emph{Medical Image Analysis}, 97:\penalty0 103280,
  2024{\natexlab{a}}.

\bibitem[Chen et~al.(2024{\natexlab{b}})Chen, Ding, Lu, Williamson, Jaume,
  Song, Chen, Zhang, Shao, Shaban, et~al.]{chen2024towards}
Richard~J Chen, Tong Ding, Ming~Y Lu, Drew~FK Williamson, Guillaume Jaume,
  Andrew~H Song, Bowen Chen, Andrew Zhang, Daniel Shao, Muhammad Shaban, et~al.
\newblock Towards a general-purpose foundation model for computational
  pathology.
\newblock \emph{Nature Medicine}, 30\penalty0 (3):\penalty0 850--862,
  2024{\natexlab{b}}.

\bibitem[Chen et~al.(2020{\natexlab{a}})Chen, Kornblith, Norouzi, and
  Hinton]{chen2020simple}
Ting Chen, Simon Kornblith, Mohammad Norouzi, and Geoffrey Hinton.
\newblock A simple framework for contrastive learning of visual
  representations.
\newblock In \emph{International conference on machine learning}, pages
  1597--1607. PMLR, 2020{\natexlab{a}}.

\bibitem[Chen and He(2021)]{Chen2021CVPR}
Xinlei Chen and Kaiming He.
\newblock Exploring simple siamese representation learning.
\newblock In \emph{Proceedings of the IEEE/CVF Conference on Computer Vision
  and Pattern Recognition (CVPR)}, pages 15750--15758, 2021.

\bibitem[Chen et~al.(2020{\natexlab{b}})Chen, Fan, Girshick, and
  He]{chen2020improved}
Xinlei Chen, Haoqi Fan, Ross Girshick, and Kaiming He.
\newblock Improved baselines with momentum contrastive learning.
\newblock \emph{arXiv preprint arXiv:2003.04297}, 2020{\natexlab{b}}.

\bibitem[Cohen et~al.(2009)Cohen, Huang, Chen, Benesty, Benesty, Chen, Huang,
  and Cohen]{cohen2009pearson}
Israel Cohen, Yiteng Huang, Jingdong Chen, Jacob Benesty, Jacob Benesty,
  Jingdong Chen, Yiteng Huang, and Israel Cohen.
\newblock Pearson correlation coefficient.
\newblock \emph{Noise reduction in speech processing}, pages 1--4, 2009.

\bibitem[Contier et~al.(2024)Contier, Baker, and
  Hebart]{contier2024distributed}
Oliver Contier, Chris~I Baker, and Martin~N Hebart.
\newblock Distributed representations of behaviour-derived object dimensions in
  the human visual system.
\newblock \emph{Nature Human Behaviour}, 8\penalty0 (11):\penalty0 2179--2193,
  2024.

\bibitem[Dong et~al.(2023)Dong, Cao, and Fu]{dong2023rethinking}
Qiaole Dong, Chenjie Cao, and Yanwei Fu.
\newblock Rethinking optical flow from geometric matching consistent
  perspective.
\newblock In \emph{Proceedings of the IEEE/CVF Conference on Computer Vision
  and Pattern Recognition}, pages 1337--1347, 2023.

\bibitem[Ehm et~al.(2024)Ehm, Gao, Roetzer, Eisenberger, Cremers, and
  Bernard]{ehm2024partial}
Viktoria Ehm, Maolin Gao, Paul Roetzer, Marvin Eisenberger, Daniel Cremers, and
  Florian Bernard.
\newblock Partial-to-partial shape matching with geometric consistency.
\newblock In \emph{Proceedings of the IEEE/CVF Conference on Computer Vision
  and Pattern Recognition}, pages 27488--27497, 2024.

\bibitem[Gonzales and Wintz(1987)]{gonzales1987digital}
Rafael~C Gonzales and Paul Wintz.
\newblock \emph{Digital image processing}.
\newblock Addison-Wesley Longman Publishing Co., Inc., 1987.

\bibitem[Grill et~al.(2020)Grill, Strub, Altch{\'e}, Tallec, Richemond,
  Buchatskaya, Doersch, Pires, Guo, Azar, et~al.]{grill2020bootstrap}
Jean-Bastien Grill, Florian Strub, Florent Altch{\'e}, Corentin Tallec, Pierre
  Richemond, Elena Buchatskaya, Carl Doersch, Bernardo Pires, Zhaohan Guo,
  Mohammad Azar, et~al.
\newblock Bootstrap your own latent: A new approach to self-supervised
  learning.
\newblock In \emph{Neural Information Processing Systems}, 2020.

\bibitem[Guttman(1945)]{guttman1945basis}
Louis Guttman.
\newblock A basis for analyzing test-retest reliability.
\newblock \emph{Psychometrika}, 10\penalty0 (4):\penalty0 255--282, 1945.

\bibitem[Haghighi et~al.(2024)Haghighi, Taher, Gotway, and
  Liang]{haghighi2024self}
Fatemeh Haghighi, Mohammad Reza~Hosseinzadeh Taher, Michael~B Gotway, and
  Jianming Liang.
\newblock Self-supervised learning for medical image analysis: Discriminative,
  restorative, or adversarial?
\newblock \emph{Medical Image Analysis}, 94:\penalty0 103086, 2024.

\bibitem[He et~al.(2020)He, Fan, Wu, Xie, and Girshick]{he2020momentum}
Kaiming He, Haoqi Fan, Yuxin Wu, Saining Xie, and Ross Girshick.
\newblock Momentum contrast for unsupervised visual representation learning.
\newblock In \emph{Proceedings of the IEEE/CVF conference on computer vision
  and pattern recognition}, pages 9729--9738, 2020.

\bibitem[He et~al.(2021{\natexlab{a}})He, Li, Ge, Yang, Kong, Zhu, Shu, Yang,
  and Li]{he2021few}
Yuting He, Tiantian Li, Rongjun Ge, Jian Yang, Youyong Kong, Jian Zhu, Huazhong
  Shu, Guanyu Yang, and Shuo Li.
\newblock Few-shot learning for deformable medical image registration with
  perception-correspondence decoupling and reverse teaching.
\newblock \emph{IEEE Journal of Biomedical and Health Informatics}, 26\penalty0
  (3):\penalty0 1177--1187, 2021{\natexlab{a}}.

\bibitem[He et~al.(2021{\natexlab{b}})He, Yang, Yang, Ge, Kong, Zhu, Zhang,
  Shao, Shu, Dillenseger, et~al.]{he2021meta}
Yuting He, Guanyu Yang, Jian Yang, Rongjun Ge, Youyong Kong, Xiaomei Zhu,
  Shaobo Zhang, Pengfei Shao, Huazhong Shu, Jean-Louis Dillenseger, et~al.
\newblock Meta grayscale adaptive network for 3d integrated renal structures
  segmentation.
\newblock \emph{Medical Image Analysis}, 71:\penalty0 102055,
  2021{\natexlab{b}}.

\bibitem[He et~al.(2022)He, Ge, Qi, Chen, Wu, Coatrieux, Yang, and
  Li]{he2022learning}
Yuting He, Rongjun Ge, Xiaoming Qi, Yang Chen, Jiasong Wu, Jean-Louis
  Coatrieux, Guanyu Yang, and Shuo Li.
\newblock Learning better registration to learn better few-shot medical image
  segmentation: Authenticity, diversity, and robustness.
\newblock \emph{IEEE Transactions on Neural Networks and Learning Systems},
  35\penalty0 (2):\penalty0 2588--2601, 2022.

\bibitem[He et~al.(2023)He, Yang, Ge, Chen, Coatrieux, Wang, and
  Li]{He_2023_CVPR}
Yuting He, Guanyu Yang, Rongjun Ge, Yang Chen, Jean-Louis Coatrieux, Boyu Wang,
  and Shuo Li.
\newblock Geometric visual similarity learning in 3d medical image
  self-supervised pre-training.
\newblock In \emph{Proceedings of the IEEE/CVF Conference on Computer Vision
  and Pattern Recognition (CVPR)}, 2023.

\bibitem[He et~al.(2024{\natexlab{a}})He, Ge, Qi, Chen, Wu, Coatrieux, Yang,
  and Li]{9842340}
Yuting He, Rongjun Ge, Xiaoming Qi, Yang Chen, Jiasong Wu, Jean-Louis
  Coatrieux, Guanyu Yang, and Shuo Li.
\newblock Learning better registration to learn better few-shot medical image
  segmentation: Authenticity, diversity, and robustness.
\newblock \emph{IEEE Transactions on Neural Networks and Learning Systems},
  35\penalty0 (2):\penalty0 2588--2601, 2024{\natexlab{a}}.

\bibitem[He et~al.(2024{\natexlab{b}})He, Huang, Jiang, Nie, Wang, Wang, and
  Chen]{he2024foundation}
Yuting He, Fuxiang Huang, Xinrui Jiang, Yuxiang Nie, Minghao Wang, Jiguang
  Wang, and Hao Chen.
\newblock Foundation model for advancing healthcare: Challenges, opportunities,
  and future directions.
\newblock \emph{arXiv preprint arXiv:2404.03264}, 2024{\natexlab{b}}.

\bibitem[He et~al.(2025)He, Wang, Ge, Chen, Yang, and
  Li]{he2025homeomorphismpriorfalsepositive}
Yuting He, Boyu Wang, Rongjun Ge, Yang Chen, Guanyu Yang, and Shuo Li.
\newblock Homeomorphism prior for false positive and negative problem in
  medical image dense contrastive representation learning, 2025.

\bibitem[Hinde and Dem{\'e}trio(1998)]{hinde1998overdispersion}
John Hinde and Clarice~GB Dem{\'e}trio.
\newblock Overdispersion: models and estimation.
\newblock \emph{Computational statistics \& data analysis}, 27\penalty0
  (2):\penalty0 151--170, 1998.

\bibitem[Hoffer and Ailon(2015)]{hoffer2015deep}
Elad Hoffer and Nir Ailon.
\newblock Deep metric learning using triplet network.
\newblock In \emph{Similarity-based pattern recognition: third international
  workshop, SIMBAD 2015, Copenhagen, Denmark, October 12-14, 2015. Proceedings
  3}, pages 84--92. Springer, 2015.

\bibitem[Hofmanninger et~al.(2020)Hofmanninger, Prayer, Pan, R{\"o}hrich,
  Prosch, and Langs]{hofmanninger2020automatic}
Johannes Hofmanninger, Forian Prayer, Jeanny Pan, Sebastian R{\"o}hrich, Helmut
  Prosch, and Georg Langs.
\newblock Automatic lung segmentation in routine imaging is primarily a data
  diversity problem, not a methodology problem.
\newblock \emph{European Radiology Experimental}, 4:\penalty0 1--13, 2020.

\bibitem[Isensee et~al.(2019)Isensee, Schell, Pflueger, Brugnara, Bonekamp,
  Neuberger, Wick, Schlemmer, Heiland, Wick, et~al.]{isensee2019automated}
Fabian Isensee, Marianne Schell, Irada Pflueger, Gianluca Brugnara, David
  Bonekamp, Ulf Neuberger, Antje Wick, Heinz-Peter Schlemmer, Sabine Heiland,
  Wolfgang Wick, et~al.
\newblock Automated brain extraction of multisequence mri using artificial
  neural networks.
\newblock \emph{Human brain mapping}, 40\penalty0 (17):\penalty0 4952--4964,
  2019.

\bibitem[Iverson(1962)]{iverson1962programming}
Kenneth~E Iverson.
\newblock A programming language.
\newblock In \emph{Proceedings of the May 1-3, 1962, spring joint computer
  conference}, pages 345--351, 1962.

\bibitem[Jin et~al.(2022)Jin, Huang, Zhou, Li, Yan, Sun, Zhang, Wang, and
  Ye]{jin2022fives}
Kai Jin, Xingru Huang, Jingxing Zhou, Yunxiang Li, Yan Yan, Yibao Sun, Qianni
  Zhang, Yaqi Wang, and Juan Ye.
\newblock Fives: A fundus image dataset for artificial intelligence based
  vessel segmentation.
\newblock \emph{Scientific data}, 9\penalty0 (1):\penalty0 475, 2022.

\bibitem[Kennedy et~al.(2012)Kennedy, Haselgrove, Hodge, Rane, Makris, and
  Frazier]{kennedy2012candishare}
David~N Kennedy, Christian Haselgrove, Steven~M Hodge, Pallavi~S Rane, Nikos
  Makris, and Jean~A Frazier.
\newblock Candishare: A resource for pediatric neuroimaging data.
\newblock \emph{Neuroinformatics}, 10\penalty0 (3):\penalty0 319, 2012.

\bibitem[Kermany et~al.(2018)Kermany, Goldbaum, Cai, Valentim, Liang, Baxter,
  McKeown, Yang, Wu, Yan, et~al.]{kermany2018identifying}
Daniel~S Kermany, Michael Goldbaum, Wenjia Cai, Carolina~CS Valentim, Huiying
  Liang, Sally~L Baxter, Alex McKeown, Ge Yang, Xiaokang Wu, Fangbing Yan,
  et~al.
\newblock Identifying medical diagnoses and treatable diseases by image-based
  deep learning.
\newblock \emph{cell}, 172\penalty0 (5):\penalty0 1122--1131, 2018.

\bibitem[Kingma and Ba(2014)]{kingma2014adam}
Diederik~P Kingma and Jimmy Ba.
\newblock Adam: A method for stochastic optimization.
\newblock \emph{arXiv preprint arXiv:1412.6980}, 2014.

\bibitem[Koltchinskii(2006)]{koltchinskii2006local}
Vladimir Koltchinskii.
\newblock Local rademacher complexities and oracle inequalities in risk
  minimization.
\newblock 2006.

\bibitem[Komodakis and Gidaris(2018)]{komodakis2018unsupervised}
Nikos Komodakis and Spyros Gidaris.
\newblock Unsupervised representation learning by predicting image rotations.
\newblock In \emph{International Conference on Learning Representations
  (ICLR)}, 2018.

\bibitem[Li et~al.(2024)Li, Liu, Li, Wang, Liu, and Yuan]{li2024ukan}
Chenxin Li, Xinyu Liu, Wuyang Li, Cheng Wang, Hengyu Liu, and Yixuan Yuan.
\newblock U-kan makes strong backbone for medical image segmentation and
  generation.
\newblock \emph{arXiv preprint}, 2024.

\bibitem[Li et~al.(2021)Li, Zhou, Zhang, Zhang, Wang, Jiang, Wu, and
  Wang]{li2021dense}
Xiaoni Li, Yu Zhou, Yifei Zhang, Aoting Zhang, Wei Wang, Ning Jiang, Haiying
  Wu, and Weiping Wang.
\newblock Dense semantic contrast for self-supervised visual representation
  learning.
\newblock In \emph{Proceedings of the 29th ACM International Conference on
  Multimedia}, pages 1368--1376, 2021.

\bibitem[Manna et~al.()Manna, Bhattacharya, and Pal]{mannaself}
Siladittya Manna, Saumik Bhattacharya, and Umapada Pal.
\newblock Self-supervised visual representation learning for medical image
  analysis: A comprehensive survey.
\newblock \emph{Transactions on Machine Learning Research}.

\bibitem[Mao et~al.(2023)Mao, Mohri, and Zhong]{mao2023cross}
Anqi Mao, Mehryar Mohri, and Yutao Zhong.
\newblock Cross-entropy loss functions: Theoretical analysis and applications.
\newblock In \emph{International conference on Machine learning}, pages
  23803--23828. PMLR, 2023.

\bibitem[Marek et~al.(2011)Marek, Jennings, Lasch, Siderowf, Tanner, Simuni,
  Coffey, Kieburtz, Flagg, Chowdhury, et~al.]{marek2011parkinson}
Kenneth Marek, Danna Jennings, Shirley Lasch, Andrew Siderowf, Caroline Tanner,
  Tanya Simuni, Chris Coffey, Karl Kieburtz, Emily Flagg, Sohini Chowdhury,
  et~al.
\newblock The parkinson progression marker initiative (ppmi).
\newblock \emph{Progress in neurobiology}, 95\penalty0 (4):\penalty0 629--635,
  2011.

\bibitem[Mohri(2018)]{mohri2018foundations}
Mehryar Mohri.
\newblock Foundations of machine learning, 2018.

\bibitem[Moor et~al.(2023)Moor, Banerjee, Abad, Krumholz, Leskovec, Topol, and
  Rajpurkar]{moor2023foundation}
Michael Moor, Oishi Banerjee, Zahra Shakeri~Hossein Abad, Harlan~M Krumholz,
  Jure Leskovec, Eric~J Topol, and Pranav Rajpurkar.
\newblock Foundation models for generalist medical artificial intelligence.
\newblock \emph{Nature}, 616\penalty0 (7956):\penalty0 259--265, 2023.

\bibitem[Netter(2014)]{netter2014atlas}
Frank~H Netter.
\newblock \emph{Atlas of human anatomy, Professional Edition E-Book: including
  NetterReference. com Access with full downloadable image Bank}.
\newblock Elsevier health sciences, 2014.

\bibitem[O~Pinheiro et~al.(2020)O~Pinheiro, Almahairi, Benmalek, Golemo, and
  Courville]{o2020unsupervised}
Pedro~O O~Pinheiro, Amjad Almahairi, Ryan Benmalek, Florian Golemo, and Aaron~C
  Courville.
\newblock Unsupervised learning of dense visual representations.
\newblock \emph{Advances in Neural Information Processing Systems},
  33:\penalty0 4489--4500, 2020.

\bibitem[Oord et~al.(2018)Oord, Li, and Vinyals]{oord2018representation}
Aaron van~den Oord, Yazhe Li, and Oriol Vinyals.
\newblock Representation learning with contrastive predictive coding.
\newblock \emph{arXiv preprint arXiv:1807.03748}, 2018.

\bibitem[Oquab et~al.(2024)Oquab, Darcet, Moutakanni, Vo, Szafraniec, Khalidov,
  Fernandez, Haziza, Massa, El-Nouby, et~al.]{oquab2024dinov2}
Maxime Oquab, Timoth{\'e}e Darcet, Th{\'e}o Moutakanni, Huy Vo, Marc
  Szafraniec, Vasil Khalidov, Pierre Fernandez, Daniel Haziza, Francisco Massa,
  Alaaeldin El-Nouby, et~al.
\newblock Dinov2: Learning robust visual features without supervision.
\newblock \emph{Transactions on Machine Learning Research Journal}, pages
  1--31, 2024.

\bibitem[Paszke et~al.(2019)Paszke, Gross, Massa, Lerer, Bradbury, Chanan,
  Killeen, Lin, Gimelshein, Antiga, et~al.]{paszke2019pytorch}
Adam Paszke, Sam Gross, Francisco Massa, Adam Lerer, James Bradbury, Gregory
  Chanan, Trevor Killeen, Zeming Lin, Natalia Gimelshein, Luca Antiga, et~al.
\newblock Pytorch: An imperative style, high-performance deep learning library.
\newblock \emph{Advances in neural information processing systems}, 32, 2019.

\bibitem[Pathak et~al.(2016)Pathak, Krahenbuhl, Donahue, Darrell, and
  Efros]{pathak2016context}
Deepak Pathak, Philipp Krahenbuhl, Jeff Donahue, Trevor Darrell, and Alexei~A
  Efros.
\newblock Context encoders: Feature learning by inpainting.
\newblock In \emph{Proceedings of the IEEE conference on computer vision and
  pattern recognition}, pages 2536--2544, 2016.

\bibitem[Payette et~al.(2023)Payette, Li, de~Dumast, Licandro, Ji, Siddiquee,
  Xu, Myronenko, Liu, Pei, et~al.]{payette2023fetal}
Kelly Payette, Hongwei~Bran Li, Priscille de Dumast, Roxane Licandro, Hui Ji,
  Md~Mahfuzur~Rahman Siddiquee, Daguang Xu, Andriy Myronenko, Hao Liu, Yuchen
  Pei, et~al.
\newblock Fetal brain tissue annotation and segmentation challenge results.
\newblock \emph{Medical image analysis}, 88:\penalty0 102833, 2023.

\bibitem[Qu et~al.(2005)Qu, Hariri, and Yousif]{qu2005new}
Guangzhi Qu, Salim Hariri, and Mazin Yousif.
\newblock A new dependency and correlation analysis for features.
\newblock \emph{IEEE Transactions on Knowledge and Data Engineering},
  17\penalty0 (9):\penalty0 1199--1207, 2005.

\bibitem[Quan et~al.(2024)Quan, Yao, Zhu, and Zhou]{quan2024igu}
Quan Quan, Qingsong Yao, Heqin Zhu, and S~Kevin Zhou.
\newblock Igu-aug: Information-guided unsupervised augmentation and pixel-wise
  contrastive learning for medical image analysis.
\newblock \emph{IEEE Transactions on Medical Imaging}, 2024.

\bibitem[Revel et~al.(2021)Revel, Boussouar, de~Margerie-Mellon, Saab, Lapotre,
  Mompoint, Chassagnon, Milon, Lederlin, Bennani, et~al.]{revel2021study}
Marie-Pierre Revel, Samia Boussouar, Constance de Margerie-Mellon, In{\`e}s
  Saab, Thibaut Lapotre, Dominique Mompoint, Guillaume Chassagnon, Audrey
  Milon, Mathieu Lederlin, Souhail Bennani, et~al.
\newblock Study of thoracic ct in covid-19: The stoic project.
\newblock \emph{Radiology}, 301\penalty0 (1):\penalty0 E361--E370, 2021.

\bibitem[Rocco et~al.(2017)Rocco, Arandjelovic, and
  Sivic]{rocco2017convolutional}
Ignacio Rocco, Relja Arandjelovic, and Josef Sivic.
\newblock Convolutional neural network architecture for geometric matching.
\newblock In \emph{Proceedings of the IEEE conference on computer vision and
  pattern recognition}, pages 6148--6157, 2017.

\bibitem[Ronneberger et~al.(2015)Ronneberger, Fischer, and
  Brox]{ronneberger2015u}
Olaf Ronneberger, Philipp Fischer, and Thomas Brox.
\newblock U-net: Convolutional networks for biomedical image segmentation.
\newblock In \emph{International Conference on Medical image computing and
  computer-assisted intervention}, pages 234--241. Springer, 2015.

\bibitem[Saunshi et~al.(2019)Saunshi, Plevrakis, Arora, Khodak, and
  Khandeparkar]{saunshi2019theoretical}
Nikunj Saunshi, Orestis Plevrakis, Sanjeev Arora, Mikhail Khodak, and
  Hrishikesh Khandeparkar.
\newblock A theoretical analysis of contrastive unsupervised representation
  learning.
\newblock In \emph{International Conference on Machine Learning}, pages
  5628--5637. PMLR, 2019.

\bibitem[Shiraishi et~al.(2000)Shiraishi, Katsuragawa, Ikezoe, Matsumoto,
  Kobayashi, Komatsu, Matsui, Fujita, Kodera, and
  Doi]{shiraishi2000development}
Junji Shiraishi, Shigehiko Katsuragawa, Junpei Ikezoe, Tsuneo Matsumoto,
  Takeshi Kobayashi, Ken-ichi Komatsu, Mitate Matsui, Hiroshi Fujita, Yoshie
  Kodera, and Kunio Doi.
\newblock Development of a digital image database for chest radiographs with
  and without a lung nodule: receiver operating characteristic analysis of
  radiologists' detection of pulmonary nodules.
\newblock \emph{American Journal of Roentgenology}, 174\penalty0 (1):\penalty0
  71--74, 2000.

\bibitem[Sohn(2016)]{sohn2016improved}
Kihyuk Sohn.
\newblock Improved deep metric learning with multi-class n-pair loss objective.
\newblock \emph{Advances in neural information processing systems}, 29, 2016.

\bibitem[Su{\'a}rez et~al.(2021)Su{\'a}rez, Garc{\'\i}a, and
  Herrera]{suarez2021tutorial}
Juan~Luis Su{\'a}rez, Salvador Garc{\'\i}a, and Francisco Herrera.
\newblock A tutorial on distance metric learning: Mathematical foundations,
  algorithms, experimental analysis, prospects and challenges.
\newblock \emph{Neurocomputing}, 425:\penalty0 300--322, 2021.

\bibitem[Taha and Hanbury(2015)]{taha2015metrics}
Abdel~Aziz Taha and Allan Hanbury.
\newblock Metrics for evaluating 3d medical image segmentation: analysis,
  selection, and tool.
\newblock \emph{BMC medical imaging}, 15\penalty0 (1):\penalty0 1--28, 2015.

\bibitem[Thoma et~al.(2020)Thoma, Paudel, and Gool]{thoma2020soft}
Janine Thoma, Danda~Pani Paudel, and Luc~V Gool.
\newblock Soft contrastive learning for visual localization.
\newblock \emph{Advances in Neural Information Processing Systems},
  33:\penalty0 11119--11130, 2020.

\bibitem[Tian et~al.(2020)Tian, Sun, Poole, Krishnan, Schmid, and
  Isola]{tian2020makes}
Yonglong Tian, Chen Sun, Ben Poole, Dilip Krishnan, Cordelia Schmid, and
  Phillip Isola.
\newblock What makes for good views for contrastive learning?
\newblock \emph{Advances in Neural Information Processing Systems},
  33:\penalty0 6827--6839, 2020.

\bibitem[Van~der Maaten and Hinton(2008)]{van2008visualizing}
Laurens Van~der Maaten and Geoffrey Hinton.
\newblock Visualizing data using t-sne.
\newblock \emph{Journal of machine learning research}, 9\penalty0 (11), 2008.

\bibitem[Van~Ginneken et~al.(2006)Van~Ginneken, Stegmann, and
  Loog]{van2006segmentation}
Bram Van~Ginneken, Mikkel~B Stegmann, and Marco Loog.
\newblock Segmentation of anatomical structures in chest radiographs using
  supervised methods: a comparative study on a public database.
\newblock \emph{Medical image analysis}, 10\penalty0 (1):\penalty0 19--40,
  2006.

\bibitem[Vaswani(2017)]{vaswani2017attention}
A Vaswani.
\newblock Attention is all you need.
\newblock \emph{Advances in Neural Information Processing Systems}, 2017.

\bibitem[Vincent et~al.(2010)Vincent, Larochelle, Lajoie, Bengio, Manzagol, and
  Bottou]{vincent2010stacked}
Pascal Vincent, Hugo Larochelle, Isabelle Lajoie, Yoshua Bengio, Pierre-Antoine
  Manzagol, and L{\'e}on Bottou.
\newblock Stacked denoising autoencoders: Learning useful representations in a
  deep network with a local denoising criterion.
\newblock \emph{Journal of machine learning research}, 11\penalty0 (12), 2010.

\bibitem[Vorontsov et~al.(2024)Vorontsov, Bozkurt, Casson, Shaikovski,
  Zelechowski, Severson, Zimmermann, Hall, Tenenholtz, Fusi,
  et~al.]{vorontsov2024foundation}
Eugene Vorontsov, Alican Bozkurt, Adam Casson, George Shaikovski, Michal
  Zelechowski, Kristen Severson, Eric Zimmermann, James Hall, Neil Tenenholtz,
  Nicolo Fusi, et~al.
\newblock A foundation model for clinical-grade computational pathology and
  rare cancers detection.
\newblock \emph{Nature medicine}, pages 1--12, 2024.

\bibitem[Wang and Liu(2021)]{wang2021understanding}
Feng Wang and Huaping Liu.
\newblock Understanding the behaviour of contrastive loss.
\newblock In \emph{Proceedings of the IEEE/CVF conference on computer vision
  and pattern recognition}, pages 2495--2504, 2021.

\bibitem[Wang et~al.(2023)Wang, Ni, and Wang]{wang2023modet}
Haiqiao Wang, Dong Ni, and Yi Wang.
\newblock Modet: Learning deformable image registration via motion
  decomposition transformer.
\newblock In \emph{International Conference on Medical Image Computing and
  Computer-Assisted Intervention}, pages 740--749. Springer, 2023.

\bibitem[Wang and Isola(2020)]{wang2020understanding}
Tongzhou Wang and Phillip Isola.
\newblock Understanding contrastive representation learning through alignment
  and uniformity on the hypersphere.
\newblock In \emph{International conference on machine learning}, pages
  9929--9939. PMLR, 2020.

\bibitem[Wang et~al.(2017)Wang, Peng, Lu, Lu, Bagheri, and
  Summers]{wang2017chestx}
Xiaosong Wang, Yifan Peng, Le Lu, Zhiyong Lu, Mohammadhadi Bagheri, and
  Ronald~M Summers.
\newblock Chestx-ray8: Hospital-scale chest x-ray database and benchmarks on
  weakly-supervised classification and localization of common thorax diseases.
\newblock In \emph{Proceedings of the IEEE conference on computer vision and
  pattern recognition}, pages 2097--2106, 2017.

\bibitem[Wang et~al.(2022{\natexlab{a}})Wang, Zhang, Shen, and
  Kong]{wang2022densecl}
Xinlong Wang, Rufeng Zhang, Chunhua Shen, and Tao Kong.
\newblock Densecl: A simple framework for self-supervised dense visual
  pre-training.
\newblock \emph{Visual Informatics}, 2022{\natexlab{a}}.

\bibitem[Wang et~al.(2022{\natexlab{b}})Wang, Li, Zhang, Wan, Zheng, Wang,
  Gong, and Liu]{wang2022exploring}
Zhaoqing Wang, Qiang Li, Guoxin Zhang, Pengfei Wan, Wen Zheng, Nannan Wang,
  Mingming Gong, and Tongliang Liu.
\newblock Exploring set similarity for dense self-supervised representation
  learning.
\newblock In \emph{Proceedings of the IEEE/CVF Conference on Computer Vision
  and Pattern Recognition}, pages 16590--16599, 2022{\natexlab{b}}.

\bibitem[Wu et~al.(2024{\natexlab{a}})Wu, Zhuang, and Chen]{wu2024large}
Linshan Wu, Jiaxin Zhuang, and Hao Chen.
\newblock Large-scale 3d medical image pre-training with geometric context
  priors.
\newblock \emph{arXiv preprint arXiv:2410.09890}, 2024{\natexlab{a}}.

\bibitem[Wu et~al.(2024{\natexlab{b}})Wu, Zhuang, and Chen]{wu2024voco}
Linshan Wu, Jiaxin Zhuang, and Hao Chen.
\newblock Voco: A simple-yet-effective volume contrastive learning framework
  for 3d medical image analysis.
\newblock In \emph{Proceedings of the IEEE/CVF Conference on Computer Vision
  and Pattern Recognition}, pages 22873--22882, 2024{\natexlab{b}}.

\bibitem[Xie et~al.(2021)Xie, Lin, Zhang, Cao, Lin, and Hu]{xie2021propagate}
Zhenda Xie, Yutong Lin, Zheng Zhang, Yue Cao, Stephen Lin, and Han Hu.
\newblock Propagate yourself: Exploring pixel-level consistency for
  unsupervised visual representation learning.
\newblock In \emph{Proceedings of the IEEE/CVF Conference on Computer Vision
  and Pattern Recognition}, pages 16684--16693, 2021.

\bibitem[Yan et~al.(2022)Yan, Cai, Jin, Miao, Guo, Harrison, Tang, Xiao, Lu,
  and Lu]{yan2022sam}
Ke Yan, Jinzheng Cai, Dakai Jin, Shun Miao, Dazhou Guo, Adam~P Harrison, Youbao
  Tang, Jing Xiao, Jingjing Lu, and Le Lu.
\newblock Sam: Self-supervised learning of pixel-wise anatomical embeddings in
  radiological images.
\newblock \emph{IEEE Transactions on Medical Imaging}, 41\penalty0
  (10):\penalty0 2658--2669, 2022.

\bibitem[Yang and Jin(2006)]{yang2006distance}
Liu Yang and Rong Jin.
\newblock Distance metric learning: A comprehensive survey.
\newblock \emph{Michigan State Universiy}, 2\penalty0 (2):\penalty0 4, 2006.

\bibitem[Zhang et~al.(2024)Zhang, Mao, Lu, Zou, Huang, Li, Li, and
  Zhang]{zhang2024single}
Yi Zhang, Yiji Mao, Xuanyu Lu, Xingyu Zou, Hao Huang, Xinyang Li, Jiayue Li,
  and Haixian Zhang.
\newblock From single to universal: tiny lesion detection in medical imaging.
\newblock \emph{Artificial Intelligence Review}, 57\penalty0 (8):\penalty0 192,
  2024.

\bibitem[Zhao et~al.(2020)Zhao, Qian, Zhang, Li, Wei, Liu, and
  Pan]{zhao2020rethinking}
Rongjian Zhao, Buyue Qian, Xianli Zhang, Yang Li, Rong Wei, Yang Liu, and
  Yinggang Pan.
\newblock Rethinking dice loss for medical image segmentation.
\newblock In \emph{2020 IEEE International Conference on Data Mining (ICDM)},
  pages 851--860. IEEE, 2020.

\bibitem[Zhou et~al.(2023)Zhou, Chia, Wagner, Ayhan, Williamson, Struyven, Liu,
  Xu, Lozano, Woodward-Court, et~al.]{zhou2023foundation}
Yukun Zhou, Mark~A Chia, Siegfried~K Wagner, Murat~S Ayhan, Dominic~J
  Williamson, Robbert~R Struyven, Timing Liu, Moucheng Xu, Mateo~G Lozano,
  Peter Woodward-Court, et~al.
\newblock A foundation model for generalizable disease detection from retinal
  images.
\newblock \emph{Nature}, 622\penalty0 (7981):\penalty0 156--163, 2023.

\bibitem[Zhou et~al.(2021)Zhou, Sodha, Pang, Gotway, and Liang]{zhou2019models}
Zongwei Zhou, Vatsal Sodha, Jiaxuan Pang, Michael~B Gotway, and Jianming Liang.
\newblock Models genesis.
\newblock \emph{Medical image analysis}, 67:\penalty0 101840, 2021.

\bibitem[Zhu et~al.(2020)Zhu, Li, Hu, Ma, Zhou, and Zheng]{zhu2020rubik}
Jiuwen Zhu, Yuexiang Li, Yifan Hu, Kai Ma, S~Kevin Zhou, and Yefeng Zheng.
\newblock Rubik’s cube+: A self-supervised feature learning framework for 3d
  medical image analysis.
\newblock \emph{Medical image analysis}, 64:\penalty0 101746, 2020.

\end{thebibliography}
}

\end{document}